\documentclass[12pt,a4paper,final]{iopart}

\usepackage{color}
\usepackage{iopams}  
\usepackage{graphicx}
\usepackage{appendix}
\usepackage{subfig}
\usepackage{graphicx}
\usepackage{color}
\usepackage{bm}
\usepackage[breaklinks=true,colorlinks=true,linkcolor=blue,urlcolor=blue,citecolor=blue]{hyperref}

\expandafter\let\csname equation*\endcsname\relax
\expandafter\let\csname endequation*\endcsname\relax

\usepackage{amsmath}
\usepackage{amsthm}
\usepackage{algorithm}
\usepackage{algpseudocode}
\usepackage{multirow}
\usepackage[super]{nth}
\usepackage{tabularx}
\usepackage[capitalise]{cleveref}
\usepackage{amssymb}
\usepackage{dsfont}
\usepackage{mathrsfs}
\usepackage{float,indentfirst}

\theoremstyle{plain}

\newtheorem{remark}{Remark}

\newtheorem{proposition}{Proposition}

\newcommand{\N}{\mathcal{N}}

\newcommand{\G}{\mathcal{G}}
\newcommand{\F}{\mathcal{F}}
\newcommand{\E}{\mathbb{E}}

\newcommand{\Cov}{\textrm{Cov}}

\newcommand{\mean}{m}
\newcommand{\pmean}{\widehat{m}}

\newcommand{\pCov}{\widehat{C}}

\newcommand{\I}{I}
\newcommand{\R}{\mathbb{R}}

\newcommand{\rhoa}{a}

\newcommand{\bpy}{\widehat{x}}

\newcommand{\bigO}{\mathcal{O}}
\newcommand{\Hess}{\nabla_{\theta}\nabla_{\theta}}

\DeclareMathOperator*{\argmin}{arg\,min}

\newcommand{\dd}{\mathrm{d}}

\newenvironment{newremark}[1]{%
    \begin{remark}#1}{%
    \Endofdef\end{remark}%
}
\newcommand{\xqed}[1]{%
    \leavevmode\unskip\penalty9999 \hbox{}\nobreak\hfill
    \quad\hbox{\ensuremath{#1}}}
\newcommand{\Endofdef}{\xqed{\lozenge}}

\definecolor{darkred}{rgb}{.6,0,0}
\definecolor{darkblue}{rgb}{0,0,.7}
\definecolor{darkgreen}{rgb}{0,.7,0}
\definecolor{darkbrown}{rgb}{0.8,0.4,0.4}
\definecolor{purple}{rgb}{0.5,0.0,0.5}
\newcommand{\dzh}[1]{{\color{black}{#1}}}

\begin{document}

\title[Gaussian Mixture Kalman Inversion]{Efficient, Multimodal, and Derivative-Free Bayesian Inference With Fisher-Rao Gradient Flows}

\author{Yifan Chen}
\address{Courant Institute, New York University, NY}
\ead{yifan.chen@nyu.edu}

\author{Daniel~Zhengyu~Huang\footnote{Author to whom any correspondence should be addressed.}}
\address{Beijing International Center for Mathematical Research, Center for Machine Learning Research,  Peking University, Beijing, China}
\ead{huangdz@bicmr.pku.edu.cn}
  
\author{Jiaoyang~Huang}
\address{University of Pennsylvania, Philadelphia, PA}
\ead{huangjy@wharton.upenn.edu}
  
\author{Sebastian Reich}
\address{Universit\"{a}t Potsdam, Potsdam, Germany}
\ead{sebastian.reich@uni-potsdam.de}
  
\author{Andrew M. Stuart}
\address{California Institute of Technology, Pasadena, CA}
\ead{astuart@caltech.edu}

\begin{abstract}

% !!!! 300 words limit
In this paper, we study efficient approximate sampling for probability distributions known up to normalization constants. We specifically focus on a problem class arising in Bayesian inference for large-scale inverse problems in science and engineering applications. The computational challenges we address with the proposed methodology are:
(i) the need for repeated evaluations of expensive forward models;
(ii) the potential existence of multiple modes; and (iii) the fact that
gradient of, or adjoint solver for, the forward model  might not be feasible.

While existing Bayesian inference methods meet some of these challenges individually, 
we propose a framework that tackles all three systematically. 
Our approach builds upon the Fisher-Rao gradient flow in probability space, yielding a dynamical system for probability densities that converges towards the target 
distribution at a uniform exponential rate. This rapid convergence is advantageous 
for the computational burden outlined in (i). We apply Gaussian mixture approximations with operator splitting techniques to simulate the flow numerically; the resulting 
approximation can capture multiple modes thus addressing (ii). 
Furthermore, we employ the Kalman methodology to facilitate a derivative-free update 
of these Gaussian components and their respective weights,  
addressing the issue in (iii).

The proposed methodology results in an efficient derivative-free posterior approximation method, flexible
enough to handle multi-modal distributions:  
Gaussian Mixture Kalman Inversion (GMKI). 
The effectiveness of GMKI is demonstrated both theoretically and numerically in several experiments with multimodal target distributions, including proof-of-concept 
and two-dimensional examples, as well as 
a large-scale application: recovering the Navier-Stokes initial condition from solution data at positive times.
\end{abstract}

\vspace{2pc}
\noindent{\it Keywords}: Bayesian Inverse Problems, Sampling, Derivative-Free Methods, Multimodal, Kalman Methodology, Fisher-Rao Gradient Flow, Gaussian Mixtures.

\submitto{\IP}
%\maketitle

% {\renewcommand{\addtocontents}[2]{} \chapter{Appendix chapter}}
% \tableofcontents

\section{Introduction}
In this paper, we introduce the posterior approximation method called Gaussian Mixture Kalman Inversion (GMKI),
designed for solution of PDE inverse problems for which forward model evaluation
is expensive, derivative/adjoint calculations cannot be used and multiple modes
are present. In subsection \ref{ssec:context} we give the context, followed
in subsection \ref{ssec:guide} with details of our guiding motivations. Subsection
\ref{ssec:methodology} describes the key ingredients of the algorithm and
subsection \ref{ssec:contribution} the contributions. In subsection
\ref{ssec:relatedwork} we give a detailed literature review and in subsection \ref{ssec:over}
we describe the organization of the paper. 
\subsection{Context}
\label{ssec:context}
Sampling a target probability distribution known up to normalization constants is a classical problem in science and engineering. 
In this paper, we focus specifically on targets resulting from
Bayesian inverse problems \cite{kaipio2006statistical,stuart2010inverse} 
involving recovery of unknown parameter 
$\theta \in \R^{N_{\theta}}$ from noisy observation $y \in \R^{N_y}$, 
through forward model
\begin{equation}
\label{eq:IPs}
    y = \G(\theta) + \eta.
\end{equation}
Here, $\G$ denotes the forward mapping which, for the problems we focus on, is 
nonlinear and requires solution of a partial differential equation (PDE) to evaluate.
The observational noise $\eta$ is here assumed to be Gaussian: $\eta \sim \N(0,\Sigma_{\eta})$. By assigning a Gaussian prior $\N(r_0, \Sigma_0)$ to the unknown $\theta$, the Bayesian 
framework leads to the posterior distribution
\begin{subequations}
\label{eq:PhiR}
\begin{align}
    \rho_{\rm post}(\theta) & \propto  \exp(-\Phi_R(\theta)),\\ 
\Phi_R(\theta) & = \Phi(\theta)+\frac{1}{2}\lVert\Sigma_{0}^{-\frac{1}{2}}(\theta - r_0) \rVert^2,\\
\Phi(\theta) & = \frac{1}{2}\lVert\Sigma_{\eta}^{-\frac{1}{2}}(y - \G(\theta)) \rVert^2.
\end{align}
\end{subequations}
Here $\Phi$ is the negative log likelihood. 
Minimization of $\Phi_R$ is a nonlinear least-squares problem which defines
the maximum a posterior (MAP) point estimator for the Bayesian inverse problem. 
It is the goal of this paper to develop an efficient method for approximating $\rho_{\rm post}$ defined
by $\Phi_R$ in the specific setting which we now outline.

\subsection{Guiding Motivations}
\label{ssec:guide}
We give more detail concerning the motivations behind the specific posterior approximation method developed
here. Firstly we note that an appropriate unit of cost in solution of Bayesian
inverse problems is the evaluation of $\G$ as this will be required multiple
times for methods such as MCMC \cite{brooks2011handbook} and SMC \cite{del2006sequential,chopin2020introduction}; when evaluation of $\G$ requires running large scale PDE solvers fast
convergence is paramount.
Secondly, we note that multiple modes, caused by multiple minimizers of $\Phi_R$, cause many methods to
become slow~\cite{tebaldi2005quantifying}, expending multiple steps in one mode before moving to another~\cite{gayrard2004metastability,gayrard2005metastability}; in addition, many Gaussian approximation based methods are unable to capture multiple modes. Nevertheless, exploring all these modes is necessary since missing one could lead to detrimental effects on engineering or science predictions;
Thirdly we note that
the gradient of $\Phi_R$ may not be available or even feasible. This might be 
because the computational models are only given as a black box (e.g. in global climate model calibration~\cite{sen2013global,schneider2017earth}), the numerical methods are not differentiable (e.g. in the embedded boundary method~\cite{peskin1977numerical,huang2018family,huang2020modeling,cao2022bayesian} and adaptive mesh refinement~\cite{berger1989local,borker2019mesh}), or because of inherently discontinuous physics (i.e., in fracture~\cite{moes1999finite} or cloud modeling~\cite{tan2018extended,lopez2022training}).
In this paper, we address these three challenges by combining, respectively,
Fisher-Rao gradient flows, Gaussian mixture approximations, and Kalman methodology. 
The resulting posterior approximation method, \textit{Gaussian Mixture Kalman Inversion} (GMKI), is fast due to the uniform exponential convergence of Fisher-Rao gradient flows, can capture multiple modes since Gaussian mixture approximations are employed, and is derivative-free thanks to the systematic Kalman methodology.

\subsection{Key Ingredients of GMKI}
\label{ssec:methodology}
In sampling, it is widely accepted practice to construct a dynamical system 
for a density that gradually evolves to the posterior distribution, or its approximation, after a specified finite time or at infinite time. Numerical approximation of this dynamics, using either particle or parametric methods, leads to practical algorithms. 
These include sequential Monte Carlo (SMC, specified finite time)
\cite{del2006sequential}, and Markov chain Monte Carlo (MCMC, infinite time)
\cite{brooks2011handbook} that are commonly used in Bayesian inference. In recent years, gradient flows in the probability space have become a popular choice of dynamical systems \cite{garcia2020bayesian,TrillosNoticeAMS,chen2023sampling}; their study
presents the opportunity to profoundly influence our understanding and development 
of sampling algorithms. 

In general, the convergence rates of different gradient flows can vary significantly. In this paper, we focus on the Fisher-Rao gradient flow of the  Kullback–Leibler (KL) divergence \cite{lu2019accelerating,lu2022birth,chen2023sampling}:
\begin{align}  
\label{eq:mean-field-Fisher-Rao-intro}
\frac{\partial \rho_t}{\partial t} 
= \rho_t\bigl(\log \rho_{\rm post} - \log \rho_t\bigr) - \rho_t\mathbb{E}_{\rho_t}[\log \rho_{\rm post} - \log \rho_t].
\end{align}
The Fisher-Rao gradient flow converges to its steady state, $\rho_{\rm post}$, exponentially fast, with a rate of $\bigO(e^{-t})$; see \cref{proposition:FR-convergence}, \cite[Theorem 4.1]{chen2023sampling}, and also \cite{lu2019accelerating,lu2022birth,domingo2023explicit}.
This convergence rate is uniform and independent of $\rho_{\rm post}$, in particular its log-Sobolev constant, which typically determines the convergence rates of other gradient flows, such as the Wasserstein gradient flow. It is worth noting that the log-Sobolev constant may behave poorly when the posterior distribution $\rho_{\rm post}$ is highly anisotropic or multimodal \cite{gayrard2004metastability,gayrard2005metastability}.
Thus, we consider \cref{eq:mean-field-Fisher-Rao-intro} as a desirable flow for sampling general distributions.

We introduce numerical approximations of~\cref{eq:mean-field-Fisher-Rao-intro} to construct practical algorithms. 
Particle methods represent the current density $\rho_t$ by a (possibly
weighted) sum of Dirac measures evaluated at an ensemble of particles. The flow \cref{eq:mean-field-Fisher-Rao-intro} can then be realized as a birth-death dynamics of these particles \cite{lu2019accelerating,lu2022birth}. However, the birth-death rate depends on the density, so it is necessary to constantly reconstruct $\rho_t$ from the empirical particle distribution.  In \cite{lu2019accelerating,lu2022birth}, kernel density estimators have been applied for the reconstruction, but their performance may be affected when the dimension of the problem becomes large. Moreover, birth-death dynamics alone cannot change the support of the distribution, so additional steps need to be added to explore the space \cite{lu2019accelerating,lu2022birth,tan2023accelerate}; such exploration steps change the dynamics and may also lead to
challenges in high dimensional problems.

Parametric methods, which reduce the gradient flow into some parametric density space, constitute another common choice of numerical approximation.
One way to do this is to project the flow \cref{eq:mean-field-Fisher-Rao-intro} into the Gaussian space \cite{chen2023gradient, chen2018natural,lambert2022variational}, via a moment closure approach. The resulting system for the mean and covariance is given by \cite{chen2023sampling}:
\begin{equation}
\begin{aligned}
\label{eq:Gaussian Fisher-Rao}
\frac{\mathrm{d} m_t}{\dd t} = C_t\E_{\rho_{\rhoa_t}}[\nabla_\theta \log \rho_{\rm post} ] \qquad 
\frac{\mathrm{d} C_t}{\dd t} = C_t + C_t \E_{\rho_{\rhoa_t}}[\Hess \log \rho_{\rm post}]C_t,
\end{aligned}
\end{equation}
where $\rho_t$ in \cref{eq:mean-field-Fisher-Rao-intro} is approximated by a Gaussian $\rho_{a_t} = \N(m_t, C_t)$ in \cref{eq:Gaussian Fisher-Rao}; 
here $a_t = (m_t, C_t)$
is the unknown parameter. We note that one may also derive the above flow by natural gradient methods in variational inference~\cite{amari1998natural,martens2020new,zhang2019fast}; see discussions in \cite{chen2023sampling}. Theoretically, it has been shown in~\cite{chen2023sampling} that \cref{eq:Gaussian Fisher-Rao} converges exponentially fast to the best Gaussian approximation of $\rho_{\rm post}$ in the KL divergence sense, when $\rho_{\rm post}$ is log-concave. 
Therefore, by simulating \cref{eq:Gaussian Fisher-Rao}, we get a Gaussian approximation of the posterior; this can be done through direct time integration or ensemble methods.
% Nevertheless, Gaussian approximation may not capture multiple modes of general $\rho_{\rm post}$. To address this issue, 

More generally, for multimodal problems, Gaussian mixture approximations have been studied in the literature under the variational inference framework~\cite{lin2019fast, lambert2022variational, huix2024theoretical}. These approaches require the evaluation of the gradient and sometimes even the Hessian matrix of $\log \rho_{\rm post}$, as shown in \cref{eq:Gaussian Fisher-Rao}, which are not directly feasible for the type of problems which are our
focus in this paper.

On the other hand, Kalman methodology has emerged as an effective methodology
for sampling for both filter and inverse problems \cite{chen2012ensemble,emerick2013investigation,iglesias2013ensemble, pathiraja2019discrete,chada2020tikhonov,schneider2020ensemble,UKI,calvello22}. Similar to the parametric methods discussed above, it relies on Gaussian approximations; however, it additionally utilizes the structure of the problem, i.e., the least-squares form of the posterior as described in \cref{eq:PhiR}. Notably, the Kalman methodology can lead to derivative-free algorithms such as the Ensemble Kalman Filter (EnKF), Unscented Kalman Filter (UKF), and Ensemble Kalman Inversion (EKI), all defined in \cite{calvello22}.
Moreover, the recent work on EKI and its variants in \cite{huang2022efficient} can be interpreted as applying Kalman-type approximations to the Fisher-Rao gradient flow \cref{eq:mean-field-Fisher-Rao-intro}, although this gradient structure was not explicitly pointed out in the original paper. The effectiveness of this method has been demonstrated on large-scale inverse problems in science and engineering, with up to hundreds of dimensions. However, since only Gaussian approximations are used, the method may not be suitable for multimodal posterior distributions.

\subsection{Contributions}
\label{ssec:contribution}
The primary focus of this paper is to extend the Kalman methodology in \cite{huang2022efficient} to Gaussian mixture approximations of the Fisher-Rao gradient flow. This leads to GMKI, a derivative-free posterior approximation method that converges fast and captures multiple modes for the challenging inversion problems studied here.
We make the following contributions:
\begin{enumerate}
    \item We propose an operator splitting approach to integrate the Fisher-Rao gradient flow in time, which leads to an exploration step that explores the space freely and an exploitation step that harnesses the data and prior information. We prove the resulting exploration-exploitation scheme converges exponentially fast to the target distribution at the discrete time level (\Cref{sec:theory}).
\item We demonstrate a connection between the continuous time limit of
the pre-existing algorithm in \cite{huang2022efficient} and Gaussian variational
inference (\Cref{sec:Kalman-Inversion-method}).
    \item We apply Gaussian mixture approximations to the exploration-exploitation scheme. We utilize the Kalman methodology to update the weights and locations of the mixtures. This leads to our derivative-free algorithm, GMKI, for sampling multimodal distributions (\Cref{sec:Gaussian-Mixture-Kalman-Inversion-method}).
    
    \item We analyze GMKI by deriving the continuous time limit of the dynamics. Based on the continuous dynamics, we study its exploration effects, establish its affine invariant property, connect the methodology to variational inference with Gaussian mixtures, and investigate the convergence properties (\Cref{sec:analysis}).
    
    \item We demonstrate, on one/two-dimensional model problems as well as a high-dimensional application (recovering the Navier-Stokes initial condition from solution data at positive times), that GMKI is able to capture multiple modes in approximately $\bigO(10)$ iterations, making it a promising approach for solving large scale Bayesian inverse problems.
    Our code is accessible online\footnote{
\tiny{\url{https://github.com/Zhengyu-Huang/InverseProblems.jl/tree/master/Multimodal}}.
} (\Cref{sec:numerics}).
\end{enumerate}

% {\color{red} Please state where in the paper each contribution can be found.}

\subsection{Literature Review} 
\label{ssec:relatedwork}

The review of relevant literature concerns SMC and MCMC, variational inference,
gradient flows and Kalman methodology. 

\subsubsection{SMC and MCMC}
Sequential Monte Carlo (SMC)
\cite{doucet2009tutorial} and Markov chain Monte Carlo (MCMC)
\cite{brooks2011handbook} are common approaches used in Bayesian inference for sampling posteriors. They lead to dynamical systems of densities that progressively converge to the target distribution. For SMC, the dynamical system operates over finite time intervals so converges fast in the density level, but numerical approximations of the dynamical system can be challenging, with difficulties such as weight collapses. Such 
issues are more pronounced in the case of multimodal posteriors, requiring a substantial
number of particles and a good initialization for SMC to succeed, due to its lack of exploration. Approximation of the finite-time dynamics in SMC via transport of measures has also been investigated \cite{klebanov2023transporting, maurais2024sampling, nikolas2024measure}. The Fisher-Rao gradient flow used in this paper can be seen as an infinite time extension of SMC dynamics that allows efficient exploration while converging exponentially fast in the density level.
MCMC approaches typically require $O(10^4)$ model runs, or more, for the type
of PDE-based inversion arising in this paper; thus they are too costly. Moreover, most MCMC approaches are based on local moves and face significant challenges in the multimodal scenario.

\subsubsection{Variational Inference} 
Variational inference \cite{jordan1999introduction,wainwright2008graphical,blei2017variational} addresses the sampling problem~\cref{eq:PhiR} using optimization, typically with a lower computational cost compared to MCMC. The objective function, often chosen to be the KL divergence between the target distribution and a variational distribution, is minimized to get a closest approximate distribution within the variational distribution family. Gaussian distributions and Gaussian mixtures are often used as the variational distribution \cite{quiroz2018gaussian,khan2017conjugate,lin2019fast, galy2021particle}. The concept of natural gradients \cite{amari1998natural,chen2018natural,lin2019fast, martens2020new,zhang2019fast} has been widely used to derive efficient optimization algorithms for variational inference. 
These algorithms typically require evaluations of gradient information for the log density. We also note that the Gaussian and Gaussian mixture ansatz has been used in conjunction with the Dirac-Frenkel variational principle to solve time-dependent PDEs of wave functions and probability densities \cite{lasser2020computing,anderson2024fisher}. When the PDE is the Fisher-Rao gradient flow, these methods can recover the parameter dynamics obtained by natural gradient flow in variational inference \cite{zhang2024sequential}.

\subsubsection{Fisher-Rao Gradient Flow}
The Fisher-Rao gradient flow plays a key role in the design of sampling algorithms studied in this paper. There is a vast literature on the use of gradient flows of the KL divergence in the density space, employing different metric tensors, for sampling. We specifically focus on the Fisher-Rao metric, introduced by C.R. Rao~\cite{rao1992information}, to derive the gradient flow~\cref{eq:mean-field-Fisher-Rao-intro}, as it is the only metric, up to scaling, invariant under any diffeomorphism of the parameter space \cite{cencov2000statistical, ay2015information, bauer2016uniqueness}. This 
invariance leads to a gradient flow converging at a rate independent of the target distribution.
In practice, the Fisher-Rao gradient flow and its simulation by birth-death processes have been used in sequential Monte Carlo samplers to reduce the variance of particle weights~\cite{del2006sequential} and accelerate Langevin sampling~\cite{lu2019accelerating,lu2022birth,tan2023accelerate} and statistical learning \cite{yan2023learning}. Kernel approximation of the flow has also been considered \cite{maurais2024sampling,wang2024measure}. Gaussian approximation of the Fisher-Rao gradient flow is studied in \cite{chen2023gradient}, with close connections to natural gradient methods in variational inference.

\subsubsection{Kalman Methodology}

The Kalman methodology encompasses a general class of approaches for solving 
filtering and inverse problems. They are based on replacing the Bayesian 
inference step in a filter, which may be viewed as governed by a prior to posterior 
map, by an approximate transport map which is exact for Gaussians; inverse
problems are solved by linking them to a filter.  Ensemble Kalman methods
give rise to derivative-free algorithms, and are appropriate for solving
filtering and inverse problems in which the desired probability distribution
is close to 
Gaussian~\cite{chen2012ensemble,ernst2015analysis,garbuno2020interacting,garbuno2020affine,huang2022efficient}.

Beyond Gaussian approximations, a strand of research has extended Kalman filters to operate on Gaussian mixtures \cite{alspach1972nonlinear, ito2000gaussian, chen2000mixture, reich2012gaussian, li2016gaussian, fan2018gaussian, grana2017bayesian, ba2022residual}. These methods model both prior and posterior distributions using Gaussian mixture distributions, leveraging a componentwise application of the Kalman methodology for each Gaussian component. 
% This effectively integrates concepts from both Kalman filtering and particle filtering in SMC.
Various techniques, such as recluster analysis and resampling techniques \cite{van2003gaussian, smith2007cluster, stordal2011bridging, hoteit2012particle, frei2013mixture}, as well as localization techniques \cite{bengtsson2003toward, sun2009sequential, stordal2012filtering}, have been developed to enhance the robustness of these approaches.
Nevertheless, existing methods in this category are tailored to transform a Gaussian mixture prior into a Gaussian mixture posterior; they can be understood as a Gaussian mixture approximation of the dynamics in SMC. The resulting methods lack full
exploration of the space of possible solutions. In contrast, 
%the GMKI method introduced in this paper is designed for sampling general multimodal posterior distributions without specifying multimodal structures on the prior. Unlike previous approaches, 
GMKI incorporates gradient flows, resulting in theoretical advantages manifest in its 
analysis. In practice, GMKI's exploration component enables effective traversal of the solution space, leading to robust performance without weight collapse.

\subsection{Organization}
\label{ssec:over}
The paper is organized as follows. 
In \Cref{sec:theory}, 
we introduce the Fisher-Rao gradient flow and the exploration-exploitation scheme for discretizing the flow in time.
In \Cref{sec:Kalman-Inversion-method}, the Gaussian approximation approach for spatial approximation is reviewed. 
In \Cref{sec:Gaussian-Mixture-Kalman-Inversion-method}, the proposed GMKI approach is presented, which relies on the Gaussian mixture approximation and Kalman methodology.
In \Cref{sec:analysis}, the continuous time dynamics of the GMKI approach is derived and analyzed.
In \Cref{sec:numerics}, numerical experiments are provided. 
We make concluding remarks in \Cref{sec:conclusion}.

\section{Fisher-Rao Gradient Flow}
\label{sec:theory}
In the following two subsections, we (i) briefly describe the Fisher-Rao gradient 
flow in the time-continuous settings; and (ii) introduce our operator splitting 
approach for simulating this flow in practice.

\subsection{Continuous Flow}
In this paper we focus on the gradient flow arising from using the KL divergence 
\begin{align}
{\rm KL}[\rho\Vert\rho_{\rm post}] = \int  \rho\log\bigl(\frac{\rho}{\rho_{\rm post}}\bigr) \dd\theta = \E_{\rho}[\log\rho - \log\rho_{\rm post}]
\end{align} 
as the energy functional, along with the Fisher-Rao metric tensor $M^{\rm FR}(\rho)^{-1}\psi = \rho \psi$; the resulting gradient flow has the form (See \cite[Section 4.1]{chen2023sampling})
\begin{align}
    \frac{\partial \rho_t}{\partial t} 
&= -M^{\rm FR}(\rho_t)^{-1} \frac{\delta {\rm KL}[\rho_t\Vert\rho_{\rm post}]}{ \delta \rho}
\nonumber\\
&=\rho_t\bigl(\log \rho_{\rm post} - \log \rho_t\bigr) - \rho_t\mathbb{E}_{\rho_t}[\log \rho_{\rm post} - \log \rho_t].\label{eqn-FR-KL}
\end{align}
We have the following uniform exponential convergence result 
for this flow (\cite[Theorem 4.1]{chen2023sampling}; see also related results in \cite{lu2019accelerating,lu2022birth,domingo2023explicit,carrillo2024fisher}):
\begin{proposition}
\label{proposition:FR-convergence}  
Let $\rho_t$ satisfy \cref{eqn-FR-KL}.  Assume there exist constants $K, B>0$ 
such that the initial density $\rho_0$ satisfies
\begin{align}\label{e:asup1}
    e^{-K(1+|\theta|^2)}\leq \frac{\rho_0(\theta)}{\rho_{\rm post}(\theta)}\leq e^{K(1+|\theta|^2)},
\end{align}
and $\rho_0, \rho_{\rm post}$ have bounded second moments
\begin{align}\label{e:asup2}
    \int |\theta|^2 \rho_0(\theta)\mathrm{d} \theta\leq B, \quad \int |\theta|^2 \rho_{\rm post}(\theta)\mathrm{d} \theta\leq B.
\end{align}
Then, for any $t\geq \log\bigl((1+B)K\bigr)$,
\begin{align}\label{e:KLconverge}
    {\rm KL}[\rho_{t} \Vert  \rho_{\rm post}]\leq (2+B+eB)Ke^{-t}.
\end{align}
\end{proposition}
Accurate numerical simulation of \cref{eqn-FR-KL} thus has the potential to
exhibit uniform exponential convergence across a wide range of targets
$\rho_{\rm post}$.

\subsection{Time-Stepping via Operator Splitting}
As a first step towards the derivation of an algorithm, we apply operator splitting 
to \cref{eqn-FR-KL}. Abusing notation we let $\rho_n$ denote our
approximation of $\rho_{t_n}$ at time $t = t_n = n\Delta t$, where $\Delta t$ denotes the time step, and
solve sequentially
\begin{equation}
\label{eqn-FR-split-1}
    \begin{aligned}
        \frac{\partial \hat{\rho}_t}{\partial t} = - \hat{\rho}_t ( \log \hat{\rho}_t  - \mathbb{E}_{\hat{\rho}_t}[\log \hat{\rho}_t]), \quad \hat{\rho}_{t_n} = \rho_{n}, \quad t_n \leq t \leq t_{n+1},
    \end{aligned}
\end{equation}
and 
\begin{equation}
\label{eqn-FR-split-2}
    \begin{aligned}
        \frac{\partial \check{\rho}_t}{\partial t} = \check{\rho}_t ( \log \rho_{\rm post}  - \mathbb{E}_{\check{\rho}_t}[\log \rho_{\rm post}]), \quad \check{\rho}_{t_n} = \hat{\rho}_{t_{n+1}}, \quad t_n \leq t \leq t_{n+1}.
    \end{aligned}
\end{equation}
The map $\rho_n \mapsto \rho_{n+1}$ is then defined by setting 
$\rho_{n+1} = \check{\rho}_{t_{n+1}}$.  
Further abusing notation we write 
$\hat{\rho}_{t_{n+1}}=\hat{\rho}_{n+1}$ and 
$\check{\rho}_{t_{n+1}}=\check{\rho}_{n+1}.$ 
Note that all of $\rho_{n},\hat{\rho}_{n+1}$ and $\check{\rho}_{n+1}$ are
functions of $\theta$.
Both \cref{eqn-FR-split-1} and \cref{eqn-FR-split-2} 
admit explicit solutions and we may write
\begin{subequations}
\begin{align}
&\hat{\rho}_{n+1}(\theta) \propto \rho_{n}(\theta)^{e^{-\Delta t}}, \label{eq-exact-split-1}\\
&\check{\rho}_{n+1}(\theta) \propto \hat{\rho}_{n+1}(\theta) \rho_{\rm post}(\theta)^{\Delta t}.\label{eq-exact-split-2}
\end{align}
\end{subequations}
Furthermore, using the first order approximation $e^{-\Delta t} \approx 1-\Delta t$
and the explicit formula \eqref{eq:PhiR} for $\rho_{\rm post},$ 
we obtain the following time-stepping scheme:
\begin{subequations}
\label{eq:concept-alg}
\begin{align}
&\hat{\rho}_{n+1}(\theta) \propto \rho_n(\theta)^{1 - \Delta t}, \label{eq:concept-alg-1}\\
&\rho_{n+1}(\theta) \propto \hat{\rho}_{n+1}(\theta) \rho_{\rm post}(\theta)^{\Delta t} \propto \hat{\rho}_{n+1}(\theta) e^{-\Delta t\Phi_R(\theta)}.\label{eq:concept-alg-2}
\end{align}
\end{subequations}

It is worth mentioning that the first order approximation corrects the bias introduced by the operator splitting, ensuring that $\rho_{\rm post}$ remains the fixed point of this time-stepping scheme. Moreover, the step \eqref{eq:concept-alg-1} and  \cref{eqn-FR-split-1} can be interpreted as the Fisher-Rao gradient flow of the negative entropy term $\E_{\rho}[\log \rho]$, which tends to increase entropy by expanding the distribution to \textit{explore} the state space. In contrast, \cref{eq:concept-alg-2}  multiplies the current distribution by the ``posterior function'' to \textit{exploit} the data and prior information, concentrating towards regions of high posterior density.
It is worth noting that the exploration-exploitation concept distinguishes the present approach from sequential Monte Carlo~\cite{del2006sequential} and other homotopy based approaches for sampling~\cite{reich2011dynamical}, which instead rely on the updating rule
$$\rho_{n+1}(\theta) \propto \rho_{n}(\theta) e^{-\Delta t\Phi(\theta)}$$
to deform the prior into the posterior in one unit time. 
The iteration \cref{eq:concept-alg} is first proposed as the basis for
sampling algorithms in \cite{huang2022efficient} as a methodology to remedy the  
ensemble collapse of EKI in long time asymptotics; however, the connection
to gradient flows is not pointed out. We note that the iteration \cref{eq:concept-alg} also connects to the tempering (or annealing) approaches that are commonly used in the Monte Carlo literature \cite{gelman1998simulating, neal2001annealed,chen2024ensemble}. Finally
\cref{eq:concept-alg} can also be interpreted as an entropic mirror descent algorithm in optimization \cite{chopin2023connection}.

The exploration-exploitation time-stepping scheme \cref{eq:concept-alg} inherits the convergence property of the continuous flow; see \cref{prop:consistency}. The proof can be found in \ref{append:convergence}.
\begin{proposition}
\label{prop:consistency}
Under the assumptions in \cref{proposition:FR-convergence}, 
let $\rho_n$ solve \cref{eq:concept-alg}, then for any $n \geq |\frac{\log((1+B)K)}{\log(1 - \Delta t)}|$, it holds that
\begin{align}
    \mathrm{KL}\left[\rho_{n} \Vert  \rho_{\rm post}\right]\leq (2+B+eB)K(1 - \Delta t)^{n}.
\end{align}
\end{proposition}

\section{Gaussian Approximation and Kalman Methodology}
\label{sec:Kalman-Inversion-method}
In this section, we discuss the Gaussian approximation of the scheme \cref{eq:concept-alg} through the Kalman methodology. In doing so we review the necessary techniques and pave the way for constructing our Gaussian mixture approximations in the next section.

In \cite{huang2022efficient}, the authors used Gaussian distributions to approximate 
the evolution of densities defined by \cref{eq:concept-alg}. 
More precisely, assume $\rho_n = \N(m_{n}, C_{n})$. Then, the first exploration step \cref{eq:concept-alg-1} leads to
\begin{align}
\label{eq:Gaussian-alg-1}
    \hat{\rho}_{n+1} = \N(\pmean_{n+1}, \pCov_{n+1}) = \N(m_{n}, \frac{1}{1 - \Delta t}C_{n}).
\end{align}
The distribution still remains Gaussian. However, the second exploitation step \cref{eq:concept-alg-2} will map out of the space of Gaussian densities, unless $\Phi_R$ 
is quadratic. In \cite{huang2022efficient}, the Kalman methodology is employed to approximate \cref{eq:concept-alg-2}, which is similar to the analysis step in the Kalman filter.
More precisely, the methodology starts with the following artificial inverse problem: 
\begin{align}
x = \F(\theta) + \nu,
\end{align}
where we have
\begin{equation}
x = \begin{bmatrix}
        y\\
        r_0
        \end{bmatrix}
\quad     
\F(\theta) = \begin{bmatrix}
        \G(\theta)\\
        \theta
        \end{bmatrix}
\quad 
\Sigma_{\nu}
    = \begin{bmatrix}
        \Sigma_{\eta} & 0\\
        0 & \Sigma_{0} 
\end{bmatrix}.
\end{equation}
Here, we set the prior on $\theta$ as $\hat{\rho}_{n+1}(\theta)$, and the observation noise $\nu \sim \N(0, \frac{1}{\Delta t}\Sigma_{\nu})$.
Following Bayes rule, the posterior distribution of the artificial inverse problem is 
\begin{align}
\label{eq:Kalman-transport}
    \rho(\theta|x) = \frac{\hat{\rho}_{n+1}(\theta)\rho(x|\theta)}{\rho(x)} \propto \hat{\rho}_{n+1}(\theta) e^{-\Delta t \Phi_R(\theta)} = \rho_{n+1}(\theta),
\end{align}
which matches the output of the step \cref{eq:concept-alg-2}. 
Here we used the fact that \cref{eq:PhiR} can be rewritten as \begin{align}
\label{eq:Phi_R_F}
\Phi_R(\theta) = \frac{1}{2}\lVert\Sigma_{\nu}^{-\frac{1}{2}}(x - \F(\theta)) \rVert^2.
\end{align}
The Kalman methodology for approximating the posterior $\rho(\theta|x)$ may
now be adopted. One first forms a Gaussian approximation of the joint distribution of $\theta$ and $\F(\theta) + \nu$, via standard moment matching, yielding
\begin{equation}
\label{eq:KF_joint}
\rho^{\rm G}(\theta, \F(\theta) + \nu) \sim \N\Big(
    \begin{bmatrix}
    \theta\\
    x
    \end{bmatrix} 
    ;
    \begin{bmatrix}
    \pmean_{n+1}\\
    \bpy_{n+1}
    \end{bmatrix}, 
    \begin{bmatrix}
  \pCov_{n+1} & \pCov^{\theta x}_{n+1}\\
  \pCov_{n+1}^{ {\theta x}^{T} } & \pCov^{xx}_{n+1}
    \end{bmatrix}
    \Big),
\end{equation}
where $\pmean_{n+1}, \pCov_{n+1}$ are as specified previously,
and 
\begin{equation}
\label{eq:KF_joint2}
\begin{split}
\bpy_{n+1} = \E[\F(\theta)], \quad
\pCov^{\theta x}_{n+1} = \mathrm{Cov}[\theta, \F(\theta)], \quad
\pCov^{xx}_{n+1} = \mathrm{Cov}[\F(\theta) + \nu] = \mathrm{Cov}[\F(\theta)] +  \frac{1}{\Delta t}\Sigma_{\nu}.
\end{split}
\end{equation}
In the above, the expectation and covariance are taken over $\theta \sim \hat{\rho}_{n+1}$. These integrals can be computed using Monte Carlo methods or quadrature rules.
% \Cref{eq:KF_joint2} can be approximated by the Monte Carlo method, which leads to ensemble Kalman filter and its variant; or special Gaussian quadrature rules, which leads to unscented Kalman filter. 

Then, one can condition the joint Gaussian distribution~\cref{eq:KF_joint} on the event $\F(\theta) + \nu = x$, to get a Gaussian approximation of the posterior $\rho(\theta|x)$. In detail, using the Gaussian conditioning formula, one obtains
\begin{subequations}
\begin{align}
        \rho_{n+1}(\theta) \approx \rho^{\rm G}(\theta | \F(\theta) + \nu = x) = \N(\theta; m_{n+1}, C_{n+1}), 
\end{align}
\end{subequations}
where 
\begin{subequations}
\label{eq:Gaussian-alg-2}
\begin{align}
        \mean_{n+1} &= \pmean_{n+1} + \pCov^{\theta x}_{n+1} (\pCov^{x x}_{n+1})^{-1} (x - \bpy_{n+1}),                   \label{eq:KF_analysis_mean}\\
         C_{n+1} &= \pCov_{n+1} - \pCov^{\theta x}_{n+1}(\pCov^{x x}_{n+1})^{-1}(\pCov_{n+1}^{ {\theta x} })^T. \label{eq:KF_analysis_cov}
\end{align}
\end{subequations}

Combining the two updates \cref{eq:Gaussian-alg-1} and \cref{eq:Gaussian-alg-2} leads to a Gaussian approximation scheme for solving the discrete Fisher-Rao gradient flow \cref{eq:concept-alg}. The scheme is based on the Kalman methodology and is derivative free. Some theoretical and numerical studies of this scheme can be found in \cite{huang2022efficient}.

\begin{newremark}
\label{remark-connecting Kalman and VI}
We can connect the Gaussian approximation based on the Kalman methodology and the approximation \cref{eq:Gaussian Fisher-Rao} obtained by Gaussian variational inference. To do so we consider the continuous time limit of \cref{eq:Gaussian-alg-2}, calculated in \cite[Eq. A.2]{huang2022efficient}, as
    \begin{equation}
\label{eq:ct-limit2}
\begin{aligned}
\frac{\dd m_t}{\dd t} &=  \pCov_t^{\theta x} \Sigma_{\nu}^{-1} (x - \bpy_t), \qquad 
\frac{\dd C_t}{\dd t}&= C_t -  \pCov_t^{\theta x}\Sigma_{\nu}^{-1} ({\pCov_t^{\theta x}})^{T}, 
\end{aligned}
\end{equation}
where 
$\bpy_t = \E_{\rho_{t}}[\F(\theta)]$, 
$\pCov_t^{\theta x}=\E_{\rho_{t}}\bigl[\bigl(\theta-m\bigr)\otimes\bigl(\F(\theta)-\E \F(\theta)\bigr)\bigr]$ and
the expectation is taken with respect to the distribution $\rho_t(\theta)= \N(\theta; m_t,C_t)$. We can view \cref{eq:ct-limit2} as a derivative-free approximation of \cref{eq:Gaussian Fisher-Rao} through the \textit{statistical linearization} \cite[Sec 4.3.2]{calvello22} approach. More precisely, by Stein's identity which utilizes the integration by parts formula for Gaussian measures, we have the relation $\E_{\rho_{t}} [\nabla_{\theta} \F(\theta)] = C_t^{-1}\pCov^{\theta x}_t$. Statistical linearization makes the approximation $\nabla_{\theta} \F(\theta) \approx C_t^{-1}\pCov^{\theta x}_t$ for all $\theta$; the approximation is exact when $\F$ is linear. Based on it, we can approximate the right hand side in the equation of the mean in \cref{eq:Gaussian Fisher-Rao} as follows:
\begin{equation}
\label{eqn:stein-identity1}
    C_t\E_{\rho_{t}}[\nabla_{\theta} \log \rho_{\rm post}] 
 = C_t\E_{\rho_{t}}[\nabla_{\theta}\F(\theta) \Sigma_{\nu}^{-1} (x - \F(\theta))] \approx \pCov_t^{\theta x} \Sigma_{\nu}^{-1} (x - \bpy_t), 
\end{equation} 
where in the first identity we used the fact $\rho_{\rm post} \propto \exp(-\frac{1}{2}\|\Sigma_\nu^{-1/2}(\F(\theta)-x)\|^2)$, and we used the statistical linearization approximation in the last derivation.

The stochastic linearization essentially approximates $\nabla_{\theta} \F(\theta)$ by a constant vector (which is why the term \textit{linearization} is used); under this approximation, the Hessian $\nabla_{\theta} \nabla_{\theta} \F(\theta)$ is zero\footnote{We can also interpret this step through the Gauss-Newton approximation.}. Based on this fact, we obtain, for the equation of the covariance in \cref{eq:Gaussian Fisher-Rao}, that

% $\nabla_{\theta}\nabla_{\theta} \log \rho_{\rm post} \approx \nabla_{\theta}\F(\theta)\Sigma_{\nu}^{-1} \nabla_{\theta}\F(\theta)^T$. Then, we get
\begin{equation}
\label{eqn:stein-identity2}
\begin{aligned}
    C_t - C_t\E_{\rho_{t}}[\nabla_{\theta}\nabla_{\theta} \log \rho_{\rm post}]C_t &\approx C_t-C_t\E_{\rho_{t}}[\nabla_{\theta}\F(\theta)\Sigma_{\nu}^{-1} \nabla_{\theta}\F(\theta)^T]C_t\\
    & \approx C_t - \pCov^{\theta x}_t\Sigma_{\nu}^{-1} ({\pCov^{\theta x}}_t)^{T}.
\end{aligned}
\end{equation}
Thus, we recover \cref{eq:ct-limit2}. In this sense, we can understand the Kalman methodology in the continuous limit as applying a statistical linearization approximation to the dynamics obtained in variational inference.
\end{newremark}

\section{Gaussian Mixture Kalman Inversion}
\label{sec:Gaussian-Mixture-Kalman-Inversion-method}
In this section, we study the use of Gaussian mixture models to approximate the 
evolution $\rho_n \mapsto \rho_{n+1}$ defined by \cref{eq:concept-alg}.
We assume that at step $n$ the distribution has the form of a $K$-component Gaussian mixture:
\begin{align*}
    \rho_n(\theta) = \sum_{k=1}^{K}w_{n,k}\N(\theta; m_{n,k}, C_{n,k}),
\end{align*}
where $w_{n,k}\geq 0$, $\sum_{k=1}^K w_{n,k} = 1$. It remains to specify
updates of the weights and Gaussian components, through the map defined by
\cref{eq:concept-alg}. In the two subsequent subsections, we consider
steps (\ref{eq:concept-alg}a) and (\ref{eq:concept-alg}b) respectively.

\subsection{The Exploration Step} The first step~\cref{eq:concept-alg-1}, $\hat{\rho}_{n+1}(\theta)$ $\propto \rho_n(\theta)^{1 - \Delta t}$, is not closed in the space of $K$-component Gaussian mixtures. Thus, we choose to approximate
$ \hat{\rho}_{n+1} $ by a new Gaussian mixture model 
$$ \hat{\rho}_{n+1} \approx   \hat{\rho}^{\rm GM}_{n+1} =\sum_{k=1}^{K}\hat{w}_{n+1,k}\N(\theta; \pmean_{n+1,k}, \pCov_{n+1,k}).$$ 
Note that $\rho_n(\theta)^{1 - \Delta t} = \rho_n(\theta)^{-\Delta t}\rho_n(\theta)$. We will determine the parameters of the new Gaussian mixture model by applying Gaussian moment matching\footnote{Such approximation along with our exploitation step will connect our GMKI to Gaussian mixture variational inference and natural gradient; see \Cref{rmk-GMKI-VI}.} to each component $\rho_n(\theta)^{-\Delta t} w_{n,k}\N(\theta; m_{n,k}, C_{n,k})$ in $\rho_n(\theta)^{-\Delta t}\rho_n(\theta)$. More specifically, we first rewrite the power of a Gaussian mixture as follows:
\begin{align*}
    \hat{\rho}_{n+1}(\theta) \propto & \Bigl(\sum_{k=1}^{K}w_{n,k}\N(\theta; m_{n,k}, C_{n,k})\Bigr)^{1 - \Delta t} \\
    =& \sum_{k=1}^{K} \Bigl[w_{n,k}\N(\theta; m_{n,k}, C_{n,k}) \Bigl(\sum_{i=1}^{K} w_{n,i}\N(\theta; m_{n,i}, C_{n,i}) \Bigr)^{-\Delta t}\Bigr] \\
    =& \sum_{k=1}^{K} \Bigl[f_{n,k}(\theta)\N(\theta; m_{n,k}, \frac{C_{n,k}}{1 - \Delta t})\Bigr],
\end{align*}
where 
$$ f_{n,k}(\theta) = \frac{(2\pi)^{\frac{\Delta t N_{\theta}}{2}}}{(1 - \Delta t)^{\frac{N_{\theta}}{2}}}
     w_{n,k}^{1 - \Delta t} {\det(C_{n,k})}^{\frac{\Delta t}{2}}
     \Bigl(\frac{w_{n,k}\N(\theta; m_{n,k}, C_{n,k})}{\sum_{i=1}^{K} w_{n,i}\N(\theta; m_{n,i}, C_{n,i})}\Bigr)^{\Delta t}.
$$
We approximate each component above by a Gaussian distribution:
\begin{align}
\label{eq:f-n-k}
    f_{n,k}(\theta)\N(\theta; m_{n,k}, \frac{C_{n,k}}{1 - \Delta t})
    \approx 
     \hat{w}_{n+1,k}\N(\theta; \pmean_{n+1,k}, \pCov_{n+1,k}),
\end{align}
where we set
\begin{subequations}
\label{eq:power-GMM}
\begin{align}
\hat{w}_{n+1,k} &= \int f_{n,k}(\theta)\N(\theta; m_{n,k}, \frac{C_{n,k}}{1 - \Delta t}) \dd\theta,\\
\pmean_{n+1,k} 
&= \frac{1}{\hat{w}_{n+1,k} }\int \theta f_{n,k}(\theta)\N(\theta; m_{n,k}, \frac{C_{n,k}}{1 - \Delta t}) \dd\theta,\\
% &= m_{n,k} + \frac{C_{n,k}}{\hat{w}_{n+1,k}(1 - \Delta t) }\int \nabla_{\theta}f_{n,k}(\theta)\N(\theta; m_{n,k}, \frac{C_{n,k}}{1 - \Delta t}) \dd\theta, \nonumber \\
\pCov_{n+1,k} &= \frac{1}{\hat{w}_{n+1,k} }\int (\theta - \pmean_{n+1,k}) (\theta - \pmean_{n+1,k})^T f_{n,k}(\theta)\N(\theta; m_{n,k}, \frac{C_{n,k}}{1 - \Delta t}) \dd\theta.
% &= \frac{C_{n,k}}{1 - \Delta t} -(\pmean_{n+1,k} - m_{n,k})(\pmean_{n+1,k} - m_{n,k})^T + \frac{C_{n,k}}{\hat{w}_{n+1,k} (1-\Delta t)^2}\int \nabla_{\theta}^2f_{n,k}(\theta)\N(\theta; m_{n,k}, \frac{C_{n,k}}{1 - \Delta t}) \dd\theta C_{n,k}. \nonumber
\label{eq:power-GMM-C}
\end{align}
\end{subequations}
Here, we determine $\hat{w}_{n+1,k} $, $\pmean_{n+1,k}$, and $\pCov_{n+1,k}$ by moment matching of both sides in \cref{eq:f-n-k}. These integrals can be evaluated by Monte Carlo method or quadrature rules. Then, we normalize $\{\hat{w}_{n+1,k}\}_{k=1}^{k = K}$ so that their summation is 1. The updates \eqref{eq:power-GMM} determine $\hat{\rho}^{\rm GM}_{n+1}$.

\subsection{The Exploitation Step}
The second step~\cref{eq:concept-alg-2} leads to
\begin{align*}
    \rho_{n+1}(\theta) 
    &\propto  \hat{\rho}^{\rm GM}_{n+1}(\theta) e^{-\Delta t \Phi_R(\theta)}\\
    &\propto \sum_{k=1}^{K} \hat{w}_{n+1,k} \N(\theta; \pmean_{n+1,k}, \pCov_{n+1,k}) e^{-\Delta t \Phi_R(\theta)}.
\end{align*}
Now, the goal is to approximate the above $\rho_{n+1}$ by a $K$-component Gaussian mixture $\sum_{k=1}^{K}w_{n+1,k}\N(\theta; m_{n+1,k}, C_{n+1,k})$. We adopt the Kalman methodology described in \Cref{sec:Kalman-Inversion-method}, to update each Gaussian component individually such that
\begin{align}
\label{eq:GMM-update}
    \hat{w}_{n+1,k} \N(\theta; \pmean_{n+1,k}, \pCov_{n+1,k}) e^{-\Delta t \Phi_R(\theta)}
    \approx 
     w_{n+1,k}\N(\theta; m_{n+1,k}, C_{n+1,k}).
\end{align}
More precisely, following \eqref{eq:Gaussian-alg-2}, for each $1\leq k \leq K$, we obtain the mean and covariance updates as
\begin{equation}
\begin{split}
\label{eq:GMM-update-2}
 m_{n+1,k} &= \pmean_{n+1,k} + \pCov_{n+1,k}^{\theta x} (\pCov_{n+1,k}^{x x})^{-1} (x - \bpy_{n+1,k}), \\
 C_{n+1,k} &= \pCov_{n+1,k} - \pCov_{n+1,k}^{\theta x}(\pCov_{n+1,k}^{x x})^{-1} (\pCov_{n+1,k}^{\theta x})^T,
\end{split}
\end{equation}
where 
\begin{equation*}
\begin{split}
\bpy_{n+1,k} =\E[\F(\theta)]
\qquad 
\pCov_{n+1,k}^{\theta x} = \mathrm{Cov}[\theta, \F(\theta)] \qquad 
\pCov_{n+1,k}^{xx} =  \mathrm{Cov}[\F(\theta)] +  \frac{1}{\Delta t}\Sigma_{\nu}, 
\end{split}
\end{equation*}
with $\theta \sim \N(\theta; \pmean_{n+1,k}, \pCov_{n+1,k})$. 
The weight $ w_{n+1,k} $ is estimated by matching \cref{eq:GMM-update} via integration
\begin{equation}
\label{eq:GMM-update-2-w}
\begin{split}
 w_{n+1,k} =  \hat{w}_{n+1,k} \int \N(\theta; \pmean_{n+1,k}, \pCov_{n+1,k}) e^{-\Delta t \Phi_R(\theta)} \dd\theta .
 % = \E[e^{-\Delta t \Phi_R(\theta)}].
\end{split}
\end{equation}

\Cref{eq:power-GMM,eq:GMM-update-2,eq:GMM-update-2-w}
define our Gaussian Mixture Kalman Inversion algorithm (GMKI), which leads to an iteration of Gaussian mixture approximations of \cref{eq:concept-alg} without using derivatives. As mentioned in \Cref{sec:Kalman-Inversion-method}, the updates involve Gaussian integration \cref{eq:power-GMM,eq:GMM-update-2,eq:GMM-update-2-w}, which can be approximated via Monte Carlo or quadrature rules. In this paper, we use the Monte Carlo method to approximate~\cref{eq:power-GMM}; these integrations do not require the forward evaluations so are inexpensive. Furthermore, we use the modified unscented transform detailed in~\cite[Definition 1]{huang2022efficient} to approximate~\cref{eq:GMM-update-2,eq:GMM-update-2-w}, which requires $(2N_{\theta} +1)K$ forward evaluations; these evaluations can be computed in parallel. The detailed algorithm is presented in \ref{sec:GMKI}.

\section{Theoretical Analysis}
\label{sec:analysis}
\begin{figure}[ht]
\centering
    \includegraphics[width=0.98\textwidth]{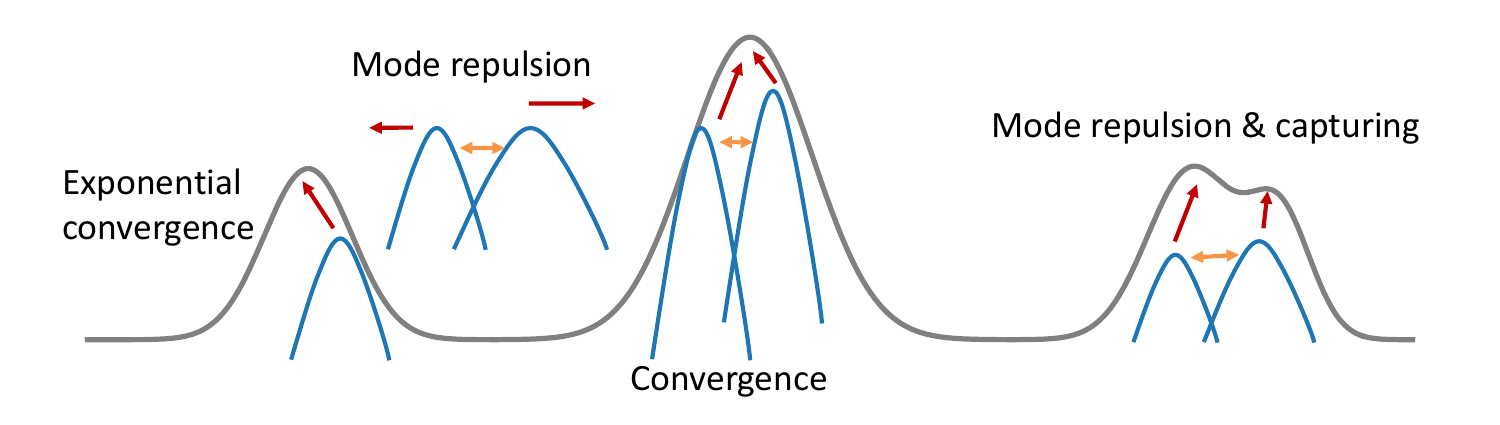}
    \caption{
    Schematic of properties of GMKI. The Grey curve represents the posterior distribution. Blue curves represent Gaussian components of the Gaussian mixture approximation. 
    From left to right: 
Gaussian components can exhibit exponential convergence toward their respective Gaussian modes if these modes are well separated (see \Cref{lem:GMKI-simplified});
the repulsion between distinct Gaussian components in the iteration of GMKI helps explore the space and capture multiple modes (see \Cref{ssec:exploration});
when multiple Gaussian components converge towards a single Gaussian mode in the posterior distribution, {they can provide a good approximation of the Gaussian mode} (see \Cref{lem:GMKI-linear});
GMKI can capture multiple modes even when these modes are intertwined (see numerical examples in \Cref{ssec:1d-bimodal}).
    }
    \label{fig:GM-Demo}
\end{figure}

In this section, theoretical studies of our GMKI methodology are presented, through a continuous time analysis.
In \Cref{ssec:exploration}, we discuss the exploration effect of the first step \cref{eq:power-GMM} of our GMKI. 
In \Cref{ssec:convergence}, we investigate the convergence properties of our GMKI method in scenarios where the posterior follows a Gaussian distribution, as well as in cases where it corresponds to a Gaussian mixture with well-separated Gaussian components. The analysis of the continuous time limit also allows us to connect GMKI with other variational inference approaches based on Gaussian mixtures. A schematic of properties of GMKI is also shown in \Cref{fig:GM-Demo}.  In \Cref{ssec:affine-invariance}, we discuss the affine invariance property of our GMKI. We summarize the conclusions
of the theoretical studies in \Cref{ssec:summary}.

\subsection{Exploration Effect}
\label{ssec:exploration}
In the derivation of GMKI, the first step \cref{eq:power-GMM} is designed to approximate the exploration phase of the Fisher-Rao gradient flow \cref{eq:concept-alg-1}. In this subsection, we investigate whether the exploration effect still persists with the approximation made by GMKI.
In fact, \cref{eq:power-GMM} tends to expand the distribution for exploration in the following two ways: (1) repulsion between Gaussian components; and (2) by an increase of the entropy.  The repulsion effect can be understood through the following continuous time limit analysis. In this analysis we abuse notation, replacing the subscript $n$ in the discrete iterations by $t$, which equals $n\Delta t$ in the continuous limit of the means, covariances and weights of the
the Gaussian mixture.
\begin{proposition}
\label{lem:GMKI-1-CTL}
The continuous time limit $(\Delta t \rightarrow 0)$ of the exploration step \cref{eq:power-GMM} is
\begin{subequations}
\label{eq:GMKI-CTL-1}
\begin{alignat}{3}
    \dot{m}_{t,k} &= -\int \N(\theta; m_{t,k}, C_{t,k}) (\theta - m_{t,k}) \log \rho_t(\theta) \dd\theta ,\label{eq:GMKI-CTL-1-m}\\
    %&=  -C_k\int \N(\theta; m_{k}, C_{k}) \nabla_{\theta} \log \rho_t(\theta) \dd\theta \nonumber \\
    \dot{C}_{t,k} 
    &= -\int \N(\theta; m_{t,k}, C_{t,k})\Bigl((\theta - m_{t,k})(\theta - m_{t,k})^{T}  - C_{t,k}\Bigr) \log \rho_t(\theta)  \dd\theta ,\\
    %&= -C_k\int \N(\theta; m_k, C_k) \nabla_{\theta}\nabla_{\theta}\log \rho_t  \dd\theta C_k \nonumber\\
    \dot{w}_{t,k}
    &= -w_{t,k}\int \Bigl[\N(\theta; m_{t,k}, C_{t,k})  - \rho_t(\theta) \Bigr] \log \rho_t(\theta) \dd\theta .
\end{alignat}
\end{subequations}
Here $\rho_t(\theta) = \sum_{k=1}^K w_{t,k} \N(\theta; m_{t,k}, C_{t,k})$.
\end{proposition}

The proof is in \ref{proof:lem:GMKI-1-CTL}.
The continuous time limit of the evolution equation of the mean \cref{eq:GMKI-CTL-1-m} suggests that, if $K \ge 2$, $m_{t,k}$ will move towards the direction where $\rho_t$ is small. Indeed, let's consider a specific scenario where $N_\theta = 1$, $K = 2$ and $m_{t,1} < m_{t,2}$, In this case, we have that
\begin{equation}
\begin{split}
    \dot{m}_{t,1} 
    &= -\int(\theta - m_{t,1}) \N(\theta; m_{t,1}, C_{t,1}) \log\rho_t(\theta) \dd\theta  
    \\
    &= -\int_{\theta > m_{t,1} \cup \theta < m_{t,1}}(\theta - m_{t,1}) \N(\theta; m_{t,1}, C_{t,1}) \log\rho_t(\theta) \dd\theta 
    \\
    &= -\int_{\theta > m_{t,1}}(\theta - m_{t,1}) \N(\theta; m_{t,1}, C_{t,1}) (\log\rho_t(\theta) - \log\rho_t(2m_{t,1} - \theta)) \dd\theta 
    \\
    & < 0,
\end{split}
\end{equation}
where the third equality results from the change of variable $\theta \rightarrow 2m_{t,1} - \theta$. And the last inequality is due to the fact that
\[ \log\rho_t(\theta) - \log\rho_t(2m_{t,1} - \theta) = \log\frac{w_{t,1} \N(\theta; m_{t,1}, C_{t,1}) + w_{t,2} \N(\theta; m_{t,2}, C_{t,2}) }{w_{t,1} \N(\theta; m_{t,1}, C_{t,1}) + w_{t,2} \N(2m_{t,1} - \theta; m_{t,2}, C_{t,2})}. \]
The right hand side is non-negative, since when $\theta > m_{t,1}$, we have $\vert m_{t,2} - 2m_{t,1} + \theta\vert =  (m_{t,2} - m_{t,1}) + (\theta - m_{t,1}) > \vert m_{t,2} - \theta \vert$ and hence $\N(2m_{t,1} - \theta; m_{t,2}, C_{t,2}) < \N(\theta; m_{t,2}, C_{t,2})$. Similarly, we can also establish that $\dot{m}_{t,2} > 0$. Hence these two Gaussian means are repulsed.
% \begin{proof}
%     $$
%     \int(\theta - m_k) \N(\theta; m_{k}, C_{k}) \log \N(\theta; m_{i}, C_{i}) \dd\theta
%     = C_k C_i^{-1}(m_i - m_k)
%     $$
% \end{proof}

We can also understand the exploration effect through the increase of the entropy; see Proposition \ref{lem:exploration-entropy}. The proof can be found in \ref{proof:lem:exploration-entropy}. 
\begin{proposition}
    \label{lem:exploration-entropy}
    The entropy of the Gaussian mixture \[\rho_t(\theta) = \sum_{k=1}^K w_{t,k}\N(\theta; m_{t,k}, C_{t,k})\] obtained from  \cref{eq:GMKI-CTL-1} is non-decreasing; indeed: 
        \begin{equation}
         \begin{split}
        \frac{\dd}{\dd t} \int -\rho_t \log \rho_t \dd\theta 
        =& \sum_{k=1}^K \Bigl(\frac{\dot{w}_{t,k}^2}{ w_{t,k}} +  w_{t,k} \dot{m}_{t,k}^TC_{t,k}^{-1}\dot{m}_{t,k} + \frac{w_{t,k}}{2}{\rm tr}[\dot{C}_{t,k}^TC_{t,k}^{-1}\dot{C}_{t,k} C_{t,k}^{-1}]\Bigr) \geq 0.
    \end{split}
    \end{equation}

\end{proposition}

\subsection{Convergence Analysis}
\label{ssec:convergence}
To provide insights for the convergence of GMKI, we consider its continuous limit in time.
Similar to \cref{eq:GMKI-CTL-1}, the continuous time limit of our GMKI is given in \cref{lem:GMKI-CTL}. The proof is in \ref{proof:lem:GMKI-CTL}.
\begin{proposition}
\label{lem:GMKI-CTL}
 The continuous time limit $(\Delta t \rightarrow 0)$ of the proposed GMKI defines the evolving Gaussian mixture measure
 $$\rho_t(\theta) = \sum_{k=1}^K w_{t,k} \N(\theta; m_{t,k}, C_{t,k})$$
 where
 \begin{subequations}
 \label{eq:GMKI-CTL}
\begin{alignat}{3}
    \dot{m}_{t,k} =& -C_{t,k}\int \N(\theta; m_{t,k}, C_{t,k}) \nabla_{\theta} \log \rho_t(\theta) \dd\theta + \pCov_{t, k}^{\theta x} \Sigma_{\nu}^{-1} (x - \bpy_{t,k}), \label{eq:GMKI-CTL-m}\\
    \dot{C}_{t,k} =& -C_{t,k}\left(\int \N(\theta; m_{t,k}, C_{t,k}) \nabla_{\theta}\nabla_{\theta}\log \rho_t(\theta)  \dd\theta\right) C_{t,k} - \pCov_{t, k}^{\theta x}\Sigma_{\nu}^{-1} {\pCov_{t,k}^{T}} \label{eq:GMKI-CTL-C},\\
    \dot{w}_{t,k} =&-w_{t,k} \int \Bigl[ \N(\theta; m_{t,k}, C_{t,k})  - \rho_t(\theta) \Bigr] \Bigl[ \log \rho_t(\theta) + \Phi_R(\theta) \Bigr]\dd\theta. \label{eq:GMKI-CTL-w}
\end{alignat}
 \end{subequations}
Here 
\begin{equation}
\label{eq:E_cov_F}
\begin{split}
\bpy_{t, k} =\E[\F(\theta)],
\qquad 
\pCov_{t, k}^{\theta x} = \mathrm{Cov}[\theta, \F(\theta)], \quad \textrm{with} \quad \theta \sim \N(m_{t,k}, C_{t,k}).
\end{split}
\end{equation}
\end{proposition}
\begin{newremark}
\label{rmk-GMKI-VI}
    We can also connect the continuous limit in \Cref{lem:GMKI-CTL} with the classical variational inference approach based on Gaussian mixtures. In fact, \cref{eq:GMKI-CTL} can be obtained by combining natural gradient methods in the Gaussian mixture context and derivative-free Kalman approximation (similar to \Cref{remark-connecting Kalman and VI}); see \ref{appendix:connect GMKI and GMVI}.
\end{newremark}

To gain insight into the convergence properties of the continuous flow, 
we first study \cref{eq:GMKI-CTL} for the Gaussian posterior case;
the proof can be found in \ref{proof:lem:GMKI-linear}.
\begin{proposition}[Linear inverse problems]
    \label{lem:GMKI-linear}
    Assume $\G(\theta) = G \cdot \theta$ is linear, and the posterior is Gaussian with the form
    $$\Phi_R(\theta) = \frac{1}{2}(\theta - m_{\rm post})^TC^{-1}_{\rm post}(\theta - m_{\rm post}). $$
    Then, the {\rm KL} divergence between the Gaussian mixture \[\rho_t(\theta) = \sum_{k=1}^K w_{t,k}\N(\theta; m_{t,k}, C_{t,k})\] obtained from  \cref{eq:GMKI-CTL} and $\rho_{\rm post}$ is non-increasing:
        \begin{equation}
         \begin{split}
        \frac{\dd}{\dd t}  \mathrm{KL}[\rho_t \Vert \rho_{\rm post}]
        =& -\sum_{k=1}^K \Bigl(\frac{\dot{w}_{t,k}^2}{ w_{t,k}} +  w_{t,k} \dot{m}_{t,k}^TC_{\rm post}^{-1}\dot{m}_{t,k} + \frac{w_{t,k}}{2}{\rm tr}[\dot{C}_{t,k}^TC_{t,k}^{-1}\dot{C}_{t,k} C_{t,k}^{-1}]\Bigr) \leq 0.
    \end{split}
    \end{equation}
    Furthermore, the mean and Fisher information matrix of stationary points $\rho_{\infty}(\theta) = \sum_k w_{\infty,k}\N(\theta; m_{\infty,k}, C_{\infty,k})$ satisfy that
    \begin{align}
        \sum_{k=1}^K w_{\infty,k} m_{\infty,k} = m_{\rm post}, \qquad \rm{FIM}[\rho_{\infty}] = \int \frac{\nabla_{\theta} \rho_{\infty}\nabla_{\theta} \rho_{\infty}^T}{\rho_{\infty}} \dd\theta  = C_{\rm post}^{-1}.
    \end{align}

    % The stationary points $w_k > 0$, $m_k = m_{\rm post}$, and $C_k = C_{\rm post}$ are marginally stable. The Jacobian matrix is
    % \begin{align}
    % \frac{\partial \dot{m}_k}{ \partial m_i} = -w_i I \quad \frac{\partial \dot{m}_k}{ \partial w_i} = 0 \quad 
    % \frac{\partial \dot{w}_k}{ \partial m_i} = 0 \quad \frac{\partial \dot{w}_k}{ \partial w_i} = 0 
    % \end{align}
    % Its eigenvectors $[\Delta m_1;\, \Delta m_2;\, \cdots ;\,\Delta m_K]$ corresponding to a zero eigenvalue satisfies $\sum_k w_k \Delta m_k = 0$.
\end{proposition}

\begin{newremark}
\Cref{lem:GMKI-linear} shows that, if the posterior is Gaussian, then the KL divergence 
of the GMKI is non-increasing in time. Furthermore, the Gaussian mixture converges to a distribution $\rho_{\infty}$ from which the correct Gaussian statistics can be extracted. Nevertheless, from our current proof, it is not yet known whether $\rho_{\infty}$ converge to $\rho_{\rm post}$. We leave this question for future study.
\end{newremark}

Finally, we provide some formal analysis for the convergence of our GMKI in scenarios where the posterior distribution is close to Gaussian mixture with the same number of components, namely
\dzh{
$$ \rho_{\rm post}(\theta) \propto \exp(-\Phi_R(\theta))  \underset{\sim}{\propto} \sum_{k=1}^K w_k^{*} \N(\theta; m_k^{*}, C_k^{*}) 
\quad \textrm{with} \inf_{1 \le k \le K}\quad w_k^{*} > 0.$$
}
For simplicity, we assume these Gaussian components are well separated. It is technical to give a precise definition of the well-separatedness of different Gaussian components; our argument here is purely formal and serves to provide insights for the behavior of GMKI.
Suppose the $k$-th Gaussian component $\N(\theta; m_k(0), C_{k}(0))$ in GMKI is close to its corresponding mode (e.g., the $k$-th mode) of $\rho_{\rm post}$ while becoming well separated from other Gaussian components. In such case, we may simplify the continuous time limit~\cref{eq:GMKI-CTL} by neglecting the interaction between different Gaussian components. The simplified continuous time dynamics and its property are presented in \cref{lem:GMKI-simplified}, with derivations in \ref{append:GMKI-simplified}. 

\begin{proposition}
\label{lem:GMKI-simplified}
Consider the simplified continuous time dynamics
 \begin{subequations}
 \label{eq:GMKI-CTL-GM}
\begin{alignat}{3}
    \dot{m}_{t,k} 
    =& \ C_{t,k} (C_k^{*})^{-1}(m_k^{*} - m_{t,k}),    \label{eq:GMKI-CTL-GM-m} 
    \\
    \dot{C}_{t,k} =&\  C_{t,k} - C_{t,k} (C_k^{*})^{-1} C_{t,k},  \label{eq:GMKI-CTL-GM-C}
    \\
    \dot{w}_{t,k} 
    =&\  w_{t,k}\bigl(\log w_k^{*} - \log w_{t,k} - \sum_{i=1}^K w_{t,i}(\log w_i^{*} - \log w_{t,i})\bigr).\label{eq:GMKI-CTL-GM-w}
\end{alignat}
 \end{subequations}
Then, $m_{t,k}$, $C_{t,k}$, and $w_{t,k}$ will converge to $m_{k}^{*}$, $C_{k}^{*}$, and $w_k^{*}$ exponentially fast.
\end{proposition}
The proof of \Cref{lem:GMKI-simplified} can be found in \ref{append:proof:GMKI-simplified}.

\subsection{Affine Invariance}
\label{ssec:affine-invariance}
Sampling methods are said to be \emph{affine invariant} if the algorithm is unchanged  under any invertible affine mapping. It is a consequence of such a property that convergence rates across all Gaussians with positive covariance 
are the same. More generally affine invariant algorithms can be highly effective for highly anisotropic distributions~\cite{goodman2010ensemble,foreman2013emcee}. This effectiveness stems from their consistent behavior across all coordinate systems related by affine transformations. Specifically, the convergence properties of these
methods can be understood by examining the optimal coordinate system, which minimizes anisotropy to the fullest extent, across all affine transformations. With this in mind, the following is of interest in relation to GMKI:

\begin{proposition}
\label{lem:GMKI-affine-invariance}
     The continuous time limit  in \Cref{lem:GMKI-CTL} is affine invariant.
     Specifically, for any invertible affine mapping $\varphi : \theta \rightarrow \widetilde\theta = A\theta + b$, and define corresponding scalar, vector, and matrix transformations
\begin{align*}
\widetilde{w}_{t,k} = w_{t,k} \quad
\widetilde{m}_{t,k} = Am_{t,k} + b \quad \widetilde{C}_{t,k} = AC_{t,k} A^T,
\end{align*}
and function transformations
\begin{align*}
\widetilde{\F}(\widetilde\theta) =  \F(A^{-1}(\widetilde\theta - b)) \quad \quad \widetilde{\Phi}_R(\widetilde\theta) =  \Phi_R(A^{-1}(\widetilde\theta - b)),
\end{align*}
\dzh{The evolution equations of $\widetilde{w}_{t,k}$, $
\widetilde{m}_{t,k}$ and $\widetilde{C}_{t,k}$ remain the same, retaining the structure of the GMKI evolutions in \cref{eq:GMKI-CTL}.}
\end{proposition} 
The proof of \Cref{lem:GMKI-affine-invariance} can be found in \ref{append:proof:GMKI-affine-invariance}.

\subsection{Summary of Theoretical Analyses}
\label{ssec:summary}

Combining insights from~\cref{lem:GMKI-1-CTL,lem:GMKI-linear,lem:GMKI-simplified,lem:GMKI-affine-invariance}, we anticipate that the repulsion between distinct  Gaussian components will enable the proposed methodology GMKI to capture possible modes of the posterior distribution more effectively. Moreover, the algorithm will converge efficiently to modes of the posterior when these modes are well separated. Finally, we note that the affine invariance property of the GMKI 
shows that it will be effective for the  approximation of certain highly anisotropic distributions.

\section{Numerical Study}
\label{sec:numerics}
In this section, we present numerical studies regarding the proposed GMKI algorithm. We focus on posterior distributions of unknown parameters or fields arising in inverse problems that may exhibit multiple modes. Three types of model problems are considered:
\begin{enumerate}
    \item A one-dimensional bimodal problem: we use this problem as a proof-of-concept example. Our result demonstrates that the convergence rate remains unchanged no matter how overlapped the two modes are. This implies the independence of the convergence rate regarding potential barriers.
    \item A two-dimensional bimodal problem: we use this benchmark problem, introduced in~\cite{garbuno2020affine,pavliotis2022derivative}, to compare GMKI with other sampling methods such as the Langevin dynamics and birth death process \cite{lu2022birth}. Our result shows that GMKI is not only more accurate but also more cost-effective 
    for this specific problem.
    \item 
    A high-dimensional bimodal problem: we consider the inverse problem of recovering the initial velocity field of the Navier-Stokes flow. The problem is designed to have a symmetry which induces two modes in the posterior.  We show that GMKI can capture both modes efficiently; this indicates GMKI's potential for addressing multimodal problems in large-scale and high-dimensional applications.
\end{enumerate}
Regarding the parameter of the GMKI algorithm (\ref{sec:GMKI}), we will specify in detail the number of mixtures $K$ in each problem. For all experiments, we use the time-step size $\Delta t = 0.5$ and adopt $J = 1000$ points for the Monte Carlo estimation of \cref{eq:power-GMM}.

\subsection{One-Dimensional Bimodal Problem}
\label{ssec:1d-bimodal}
We first consider the following 1D bimodal inverse problem, associated with a forward model
$$y = \G(\theta) + \eta \quad \textrm { with } \quad y = 1, \quad \G(\theta) = \theta^2.$$
We assume the prior is $\rho_{\rm prior} \sim \N(3, 2^2)$ and consider different noise levels:
\begin{align*}
\textrm{Case A:} \quad  \eta \sim \N(0, 0.2^2);  \\
\textrm{Case B:} \quad  \eta \sim \N(0, 0.5^2); \\
\textrm{Case C:} \quad  \eta \sim \N(0, 1.0^2); \\
\textrm{Case D:} \quad  \eta \sim \N(0, 1.5^2).
\end{align*}
Note that the overlap between these two modes is larger when the noise strength is larger. For case A, the two modes are well separated, and for case D, the two modes are nearly mingled.

\begin{figure}[ht]
\centering
    \includegraphics[width=0.95\textwidth]{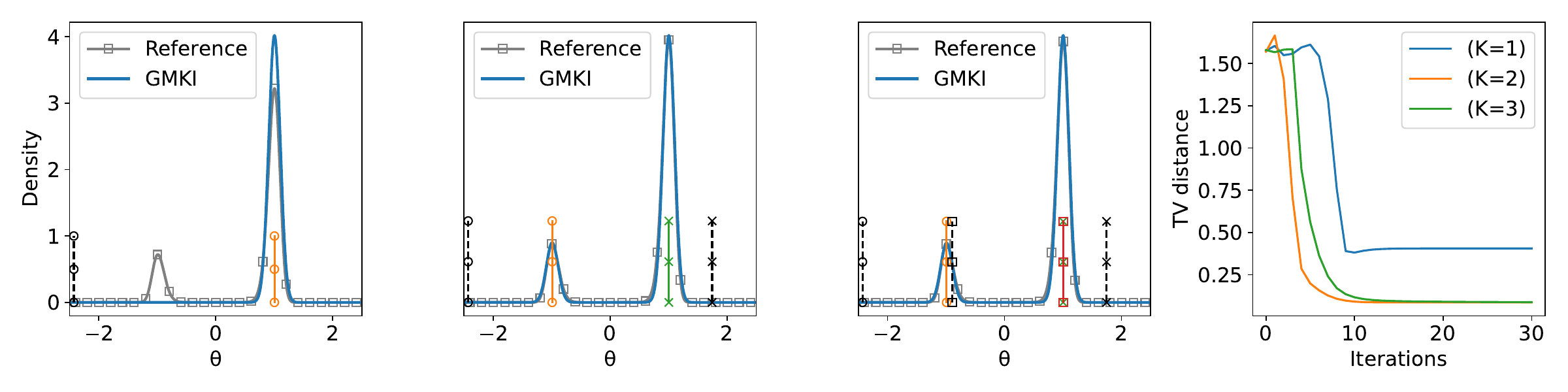}
    \includegraphics[width=0.95\textwidth]{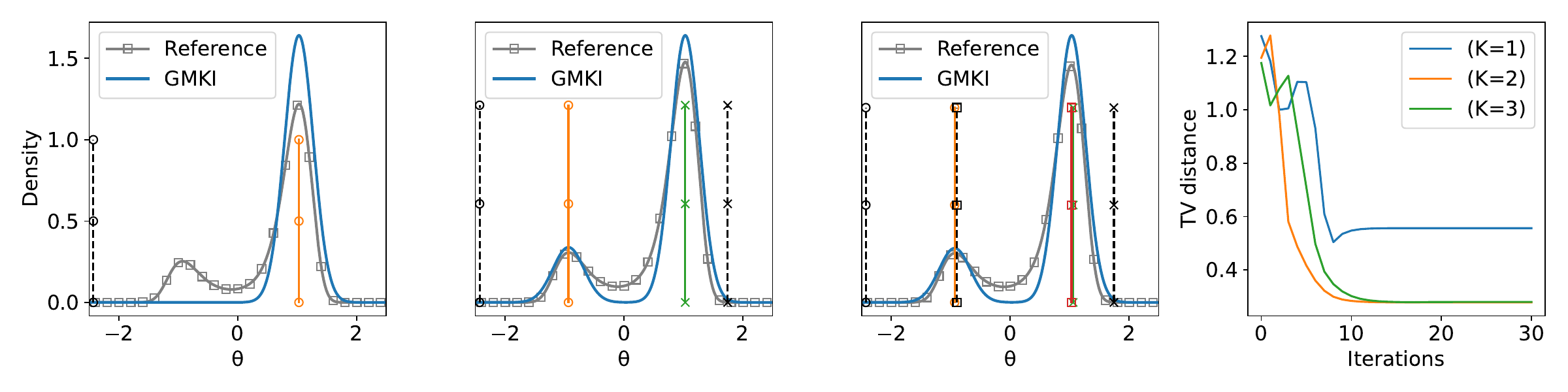}
    \includegraphics[width=0.95\textwidth]{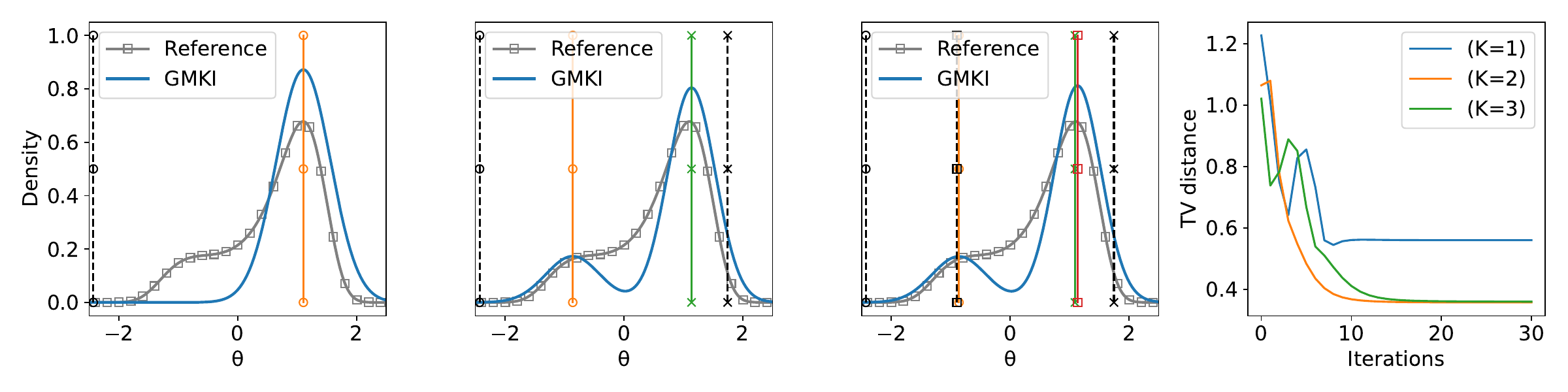}
    \includegraphics[width=0.95\textwidth]{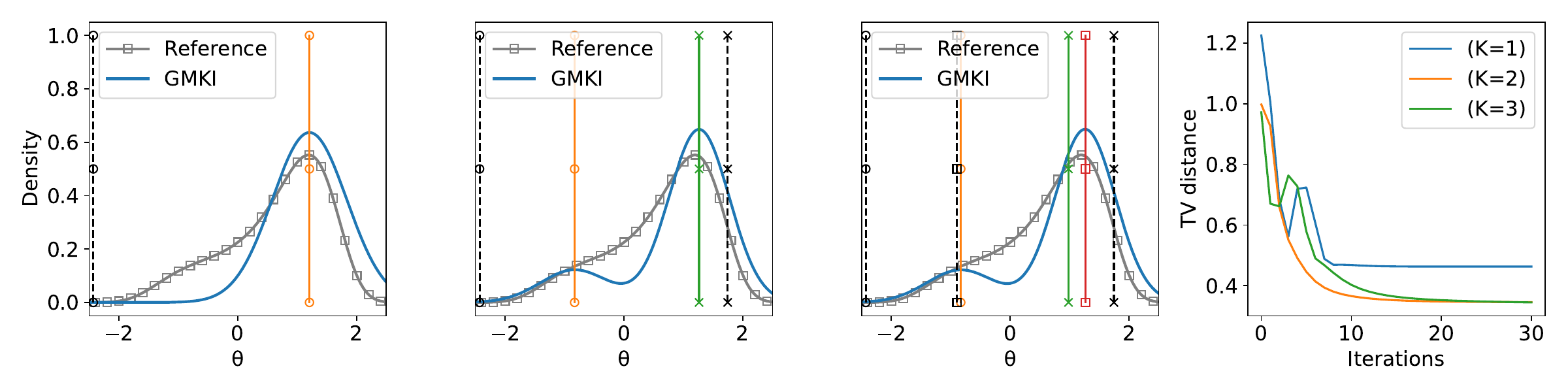}
    \caption{
    The one-dimensional bimodal problem with $\Sigma_{\eta}$ values of $0.2^2$~(top), $0.5^2$~(top middle), $1.0^2$~(bottom middle), and $1.5^2$~(bottom). 
    Each panel displays the reference posterior distribution (grey square lines) and posterior distributions estimated by the GMKI (blue lines) at the $30$th iteration with mode number $K=1,\,2,\,3$~(from left to right) with mean $m_k$ (colored) and initial mean (black) of each Gaussian component marked.
    The fourth figure shows the total variation distance between the reference posterior distribution and the posterior distributions estimated by the GMKI with mode number $K=1,\,2,\,3$.
    }
    \label{fig:1D-density-all}
\end{figure}

We apply GMKI with $K = 1$, $2$, and $3$ modes, which are randomly initialized according to the prior distribution; we assign them equal weights. In each iteration, the GMKI algorithm requires $3$, $6$, and $9$ forward evaluations, respectively. The reference posterior distribution is obtained by evaluating the unnormalized posterior on a uniform grid and then normalizing it.

The results for different cases are reported from \cref{fig:1D-density-all}.
Each row first shows the reference posterior and posteriors approximated by GMKI at the $30$th iteration,  using different mode numbers  $K=1,\,2,\,3$ from left to right.
And the fourth figure shows the convergence in terms of the total variation (TV) distance.
When $K=1$, we can only capture one mode, and this of course will not be weighted correctly since it
will have weight one by construction; when $K=2$ or $3$, we can capture both modes. It is worth mentioning that GMKI converges in fewer than 30 iterations. The convergence behavior appears independent of the potential barrier.
For case A, where the two modes are well separated, the approximated posteriors by GMKI with $K=2$ or $3$ match very well with the reference. This observation justifies our formal analysis in \cref{lem:GMKI-simplified}. For cases B,C,D, we observe that GMKI is capable of capturing both modes, however, some discrepancy will arise in the region where the modes overlap. These discrepancies persist when increasing the mode number in GMKI from $K=2$ to $K=3$.

\subsection{Two-Dimensional Bimodal Problem}
In this subsection, we consider the 2D bimodal inverse problem from~\cite{reich2021fokker,pavliotis2022derivative}, associated with the forward model
$$y = \G(\theta) + \eta \quad \textrm { with } \quad y = 4.2297, \quad \G(\theta) = (\theta_{(1)} - \theta_{(2)})^2.$$
Here $\theta = [\theta_{(1)}, \theta_{(2)}]^T$. We assume the noise distribution is $\eta \sim \N(0, I)$ and consider two different prior distributions:
\begin{align*}
&\textrm{Case A:} \quad  \rho_{\rm prior} \sim \N(0, I);  \\
&\textrm{Case B:} \quad  \rho_{\rm prior} \sim \N([0.5,0]^T, I).
\end{align*}
For case A, the two modes are symmetric with respect to the line $\theta_{(1)} - \theta_{(2)} = 0$, while for case B, the two modes are not symmetric.

\begin{figure}[ht]
\centering
    \includegraphics[width=0.98\textwidth]{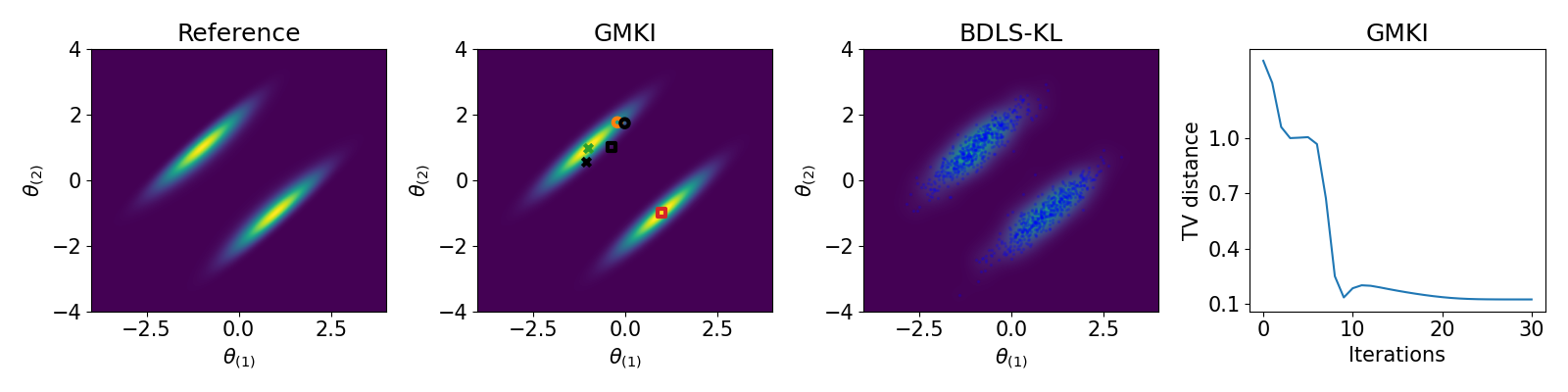}
   \caption{Two-dimensional bimodal problem with $\rho_{\rm prior} \sim \N(0, I)$. 
    From left to right: reference posterior distribution (left), posterior distributions estimated by 3-modal GMKI (middle left) at the 30th iteration (means $m_k$ (colored) and initial means (black) are marked), BDLS-KL~\cite{lu2022birth}~(middle right) at the 1000th iteration, and total variation distance between the reference posterior distribution and the posterior distributions estimated by the GMKI (right). }
    \label{fig:2D-density-0.0}
\end{figure}

\begin{figure}[ht]
\centering
    \includegraphics[width=0.98\textwidth]{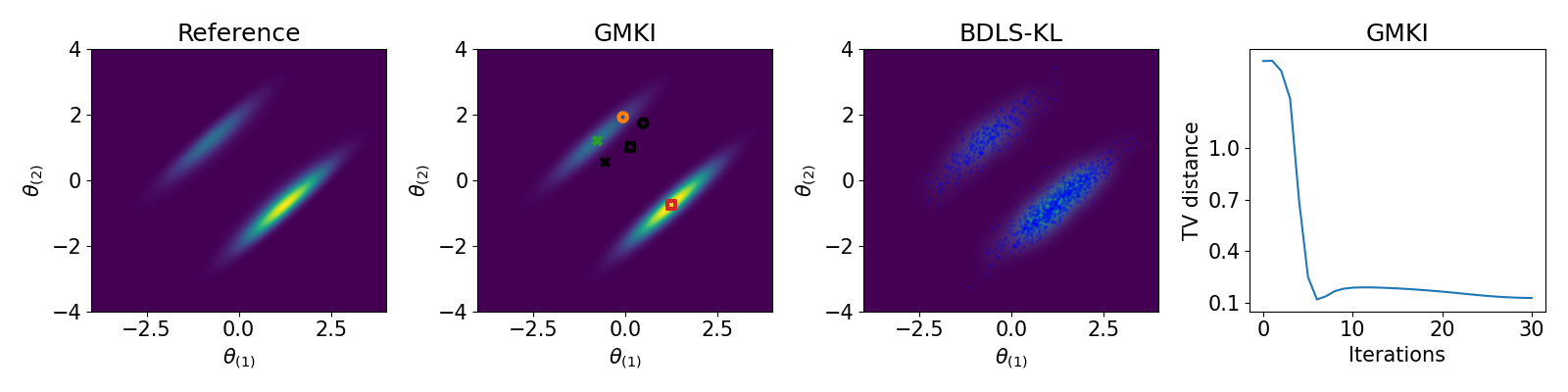}
   \caption{Two-dimensional bimodal problem with $\rho_{\rm prior} \sim \N([0.5, 0]^T, I)$. 
    From left to right: reference posterior distribution (left), posterior distributions estimated by 3-modal GMKI (middle left) at the 30th iteration (means $m_k$ (colored) and initial means (black) are marked), BDLS-KL~\cite{lu2022birth}~(middle right) at the 1000th iteration, and total variation distance between the reference posterior distribution and the posterior distributions estimated by the GMKI (right).}
    \label{fig:2D-density-0.5}
\end{figure}

%%%%%%%%%%
We apply GMKI with $K=3$ modes, which are randomly initialized based on the prior distribution and we assign these components with equal weights. 
In each iteration, the algorithm requires $(2N_{\theta}+1 )K= 15$ forward evaluations.
The reference posterior distribution is obtained by evaluating the unnormalized posterior on a uniform grid and then normalizing it.
For both case A and case B, the estimated posterior distributions obtained by the GMKI are presented in \cref{fig:2D-density-0.0,fig:2D-density-0.5}. We observe a strong correspondence with the reference, where in GMKI, mode 1 converges to one target mode, while mode 2 and mode 3 converge to another target mode. Moreover, the evolution of the total variation distance indicates rapid convergence.

As a comparison,  the derivative-free affine-invariant Langevin dynamics~(dfALDI) \cite{garbuno2020interacting,garbuno2020affine} and derivative-free Bayesian inversion using multiscale dynamics \cite{pavliotis2022derivative}  with $10^6$ iterations have been used for sampling the posterior distribution for case A. Their results are reported in \cite[Figure 5]{pavliotis2022derivative}, where dfALDI fails while multiscale dynamics can capture both modes but with wrong weights. The local preconditioner, which involves employing local empirical covariance with distance-dependent weights as introduced in \cite{reich2021fokker}, gives rise to an alternative derivative-free affine-invariant Langevin dynamics sampling approach. That approach notably enhances the sampling results in these scenarios.
Moreover, for both case A and case B, we apply the BDLS-KL algorithm proposed in \cite{lu2022birth}, which is a gradient-based sampler relying on the birth-death dynamics with kernel density estimators to approximate the Wasserstein-Fisher-Rao gradient flow. For the implementation of BDLS-KL, we use $\Delta t = 10^{-2}$, $T=10$, ensemble size $J=10^3$, and the RBF kernel $k(x,x') = \exp(\frac{1}{h}\lVert x - x'\rVert^2)$ with the bandwidth $h=\mathrm{med}^2/\log J$ adopted from \cite{liu2016stein}; here $\textrm{med}^2$ is the squared median of the pairwise Euclidean distance between the current particles. The results are shown in \cref{fig:2D-density-0.0,fig:2D-density-0.5}.  For both cases, GMKI is not only more cost-effective but also more accurate compared to these existing approaches. Moreover, to study the behavior of GMKI in higher-modal problems, a two-dimensional four-modal problem is presented in \ref{appendix-Two-Dimensional-Four-modal}, leading to a similar conclusion.

% \begin{center}
% \begin{tabular}{ c|c|c| } 
%  Rel. $L_1$ error $\frac{\lVert\rho - \rho_{\rm post} \rVert_1}{\lVert\rho_{\rm post} \rVert_1}$ & case A & case B \\ 
%   \hline
%  GMKI & cell5 & cell6 \\ 
%  WFR & cell8 & cell9 \\ 
% \end{tabular}
% \end{center}

\subsection{High-Dimensional Bimodal Problem: Navier Stokes Problem}
Finally, we study the problem of recovering the initial vorticity field $\omega_0$ of a fluid flow from measurements at later times.
The flow is described by the 2D Navier-Stokes equation on a periodic domain $D = [0,2\pi]\times[0,2\pi]$, which can be written in the vorticity-streamfunction~$\omega-\psi$ formulation:
\begin{equation}
\label{eq:NS}
\begin{split}
    &\frac{\partial \omega}{\partial t} + (v\cdot\nabla)\omega - \nu\Delta\omega = \nabla \times f, \\
    &\omega = -\Delta\psi \qquad \frac{1}{4\pi^2}\int\psi = 0
    \qquad v = [\frac{\partial \psi}{\partial x_2}, -\frac{\partial \psi}{\partial x_1}]^T + v_b.
\end{split}
\end{equation}
Here $v$ denotes the velocity vector, $\nu=0.01$ denotes the viscosity, $v_b = [0, 2\pi]^T$ denotes the non-zero mean background velocity, and $f(x) = [0,\cos(4x_{(1)})]^T$ denotes the external forcing.
%and solved by the pseudo-spectral method~\cite{hesthaven2007spectral} on a $128\times128$ grid. To eliminate aliasing error, the Orszag 2/3-Rule~\cite{orszag1972numerical} is applied and, therefore there are $85^2$ Fourier modes (padding with zeros). Time-integration is performed using the Crank–Nicolson method with $\Delta T=2.5\times 10^{-4}$.

\begin{figure}[ht]
\centering
    \includegraphics[width=0.48\textwidth]{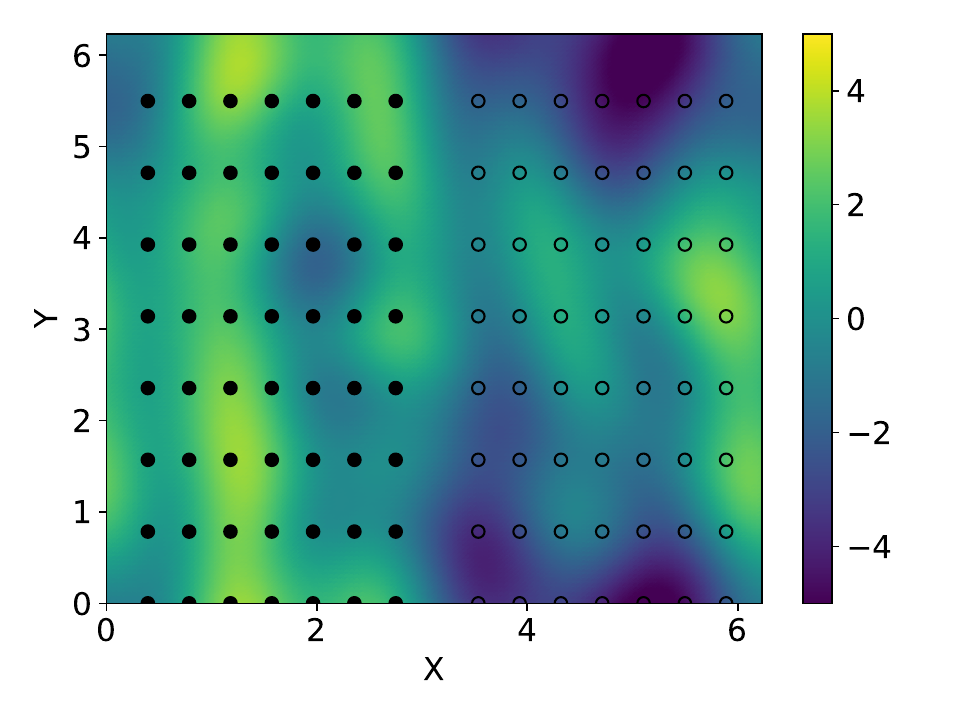}
    \includegraphics[width=0.48\textwidth]{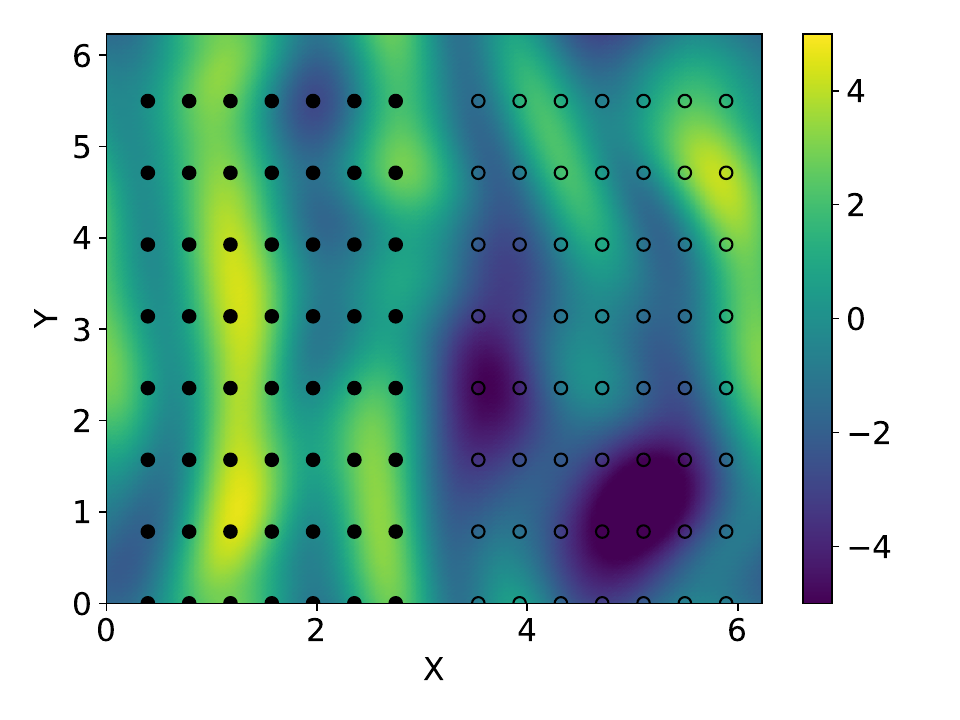}
    \caption{The vorticity field $\omega$ at $T=0.25$ and $T=0.5$ and observations $\omega([x_{(1)}, x_{(2)}]^T) - \omega([2\pi - x_{(1)}, x_{(2)}]^T)$ at $56$ equidistant points (solid black dots). Their mirroring points are marked (empty black dots).}
    \label{fig:NS-2d-ref}
\end{figure}

The problem is built to be spatially symmetric with respect to $x_{(1)} = \pi$. The source of the fluid is chosen such that 
\begin{align*}
    \nabla \times f([x_{(1)},x_{(2)}]^T) = - \nabla \times f([2\pi-x_{(1)},x_{(2)}]^T). 
\end{align*} 
The observations in the inverse problem are chosen as the difference of pointwise measurements of the vorticity value $\omega(\cdot)$
$$\omega([x_{(1)}, x_{(2)}]^T) - \omega([2\pi - x_{(1)}, x_{(2)}]^T)$$
at $56$ equidistant points in the left domain~(see \Cref{fig:NS-2d-ref}), at $T=0.25$ and $T=0.5$, corrupted with observation error $\eta \sim \N(0, 0.1^2\I)$.  
Under this set-up, both $\omega_0([x_{(1)},x_{(2)}]^T)$ and $-\omega_0([2\pi-x_{(1)},x_{(2)}]^T)$ will lead to the same measurements. Thus the inverse problem will be 
at least bi-modal.

We assume the prior of $\omega_0(x, \theta)$ is a Gaussian field with covariance operator $C = (-\Delta)^{-2}$, subject to periodic boundary conditions, on the space of mean zero functions. The corresponding KL expansion of the initial vorticity field is given by 
\begin{equation}
\label{eq:NS-KL-2d}
\omega_0(x, \theta) = \sum_{l\in K} \theta^{c}_{(l)} \sqrt{\lambda_{l}} \psi^c_l(x)  +  \theta^{s}_{(l)}\sqrt{\lambda_{l}} \psi^s_l(x),
\end{equation}
where $\mathbb{L} = \{(l_x, l_y)| l_x + l_y > 0 \textrm{ or } (l_x + l_y = 0 \textrm{ and } l_x > 0)\}$, and the eigenpairs are of the form
\begin{equation*}
    \psi^c_l(x) =\frac{\cos(l\cdot x)}{\sqrt{2}\pi}\quad \psi^s_l(x) =\frac{\sin(l\cdot x)}{\sqrt{2}\pi} \quad \lambda_l = \frac{1}{|l|^{4}},
\end{equation*}
and $\theta^{c}_{(l)},\theta^{s}_{(l)}  \sim \N(0,2\pi^2)$. The KL expansion~\cref{eq:NS-KL-2d} can be rewritten as a sum over positive integers rather than a lattice: 
\begin{equation}
\label{eq:NS-KL-1d}
    \omega_0(x,\theta) = \sum_{l=1}^{\infty} \theta_{(l)}\sqrt{\lambda_l} \psi_l(x),
\end{equation}
where the eigenvalues $\lambda_l$ are in descending order. 
% The associated forward model is
% $$y = \G(\theta) + \eta \quad \textrm { with } \quad \theta \sim \N(0, 2\pi^2I) \quad \eta \sim \N(0, 0.1^2I)$$
We truncate the expansion to the first $128$ terms and generate the true vorticity field $\omega_0(x; \theta_{\rm ref})$ with $\theta_{\rm ref} \in \R^{128}$; we aim to recover the parameter based on observation data.

% The Bayesian inference problem becomes
% \begin{align*}
% \rho_{\rm post} \propto e^{-\Phi_R(\theta)} \qquad \Phi_R(\theta) = -\frac{1}{2}(x - \F(\theta))^T\Sigma_{\nu}^{-1}(x - \F(\theta)),
% \end{align*}
% where 
% \begin{align*}
%     x = 
%     \begin{bmatrix}
%     y \\
%     m_{\rm prior}
%     \end{bmatrix}
% \quad 
%     \F = \begin{bmatrix}
%     \G(\theta) \\
%     \theta
%     \end{bmatrix}
% \quad
% \textrm{and} 
% \quad
%     \Sigma_{\nu}= \begin{bmatrix}
%     \Sigma_{\eta} & \\
%           & \Sigma_{\rm prior}
%     \end{bmatrix}
% \end{align*}

\begin{figure}[ht]
\centering
    \includegraphics[width=0.98\textwidth]{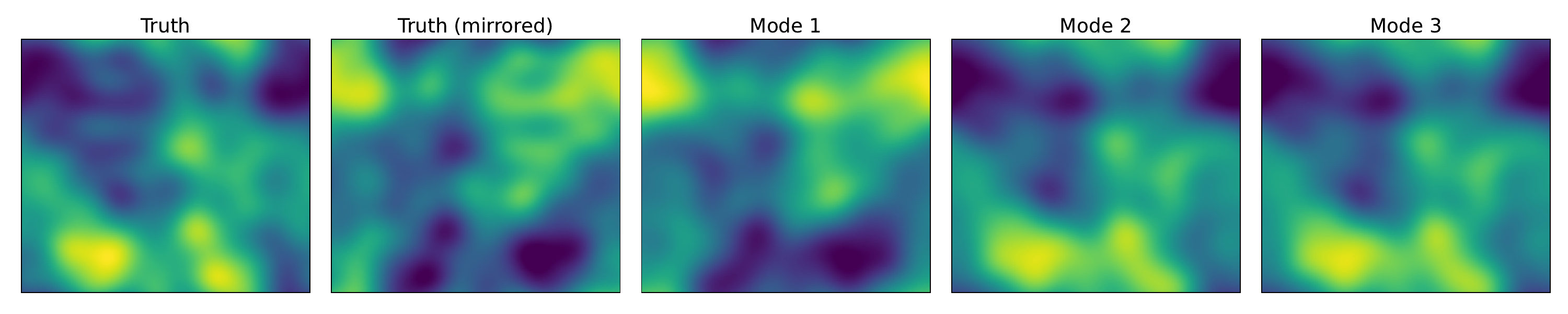}
    \caption{The true initial vorticity field $ \omega_0(x; \theta_{\rm ref})$, and recovered initial vorticity fields $\omega_0(x; m_k)$ obtained by GMKI.}
    \label{fig:NS-2D-vor}
\end{figure}

\begin{figure}[ht]
\centering
    \includegraphics[width=0.98\textwidth]{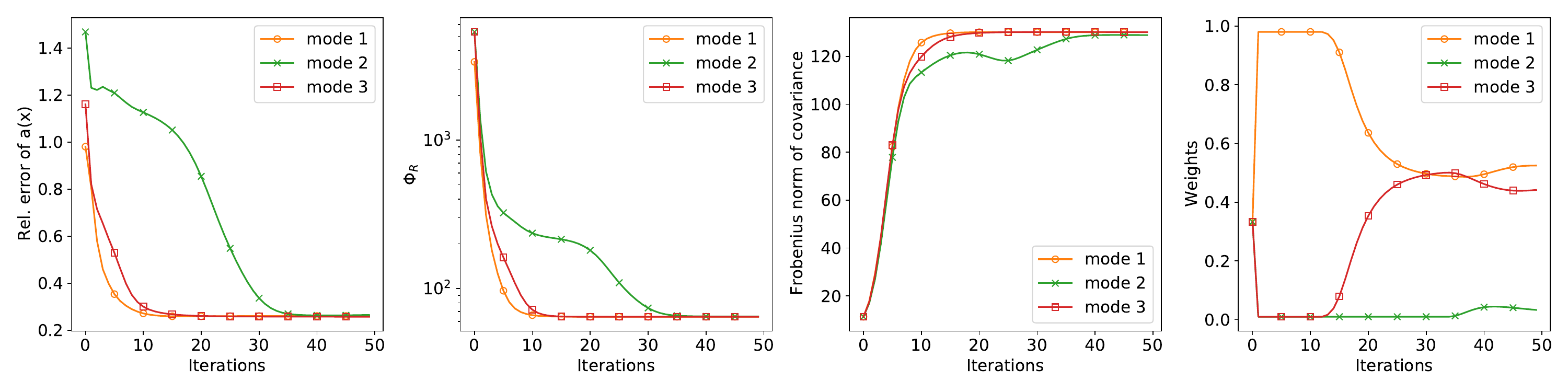}
    \caption{Navier-Stokes flow problem: the relative errors of the initial vorticity field, the optimization errors $\Phi_R(m_{n,k})$, the Frobenius norm $\lVert C_{n,k}\rVert_F$, and the Gaussian mixture weights $w_{n,k}$ (from left to right) for different modes.}
    \label{fig:NS-2D-convergence}
\end{figure}

\begin{figure}[ht]
\centering
    \includegraphics[width=0.98\textwidth]{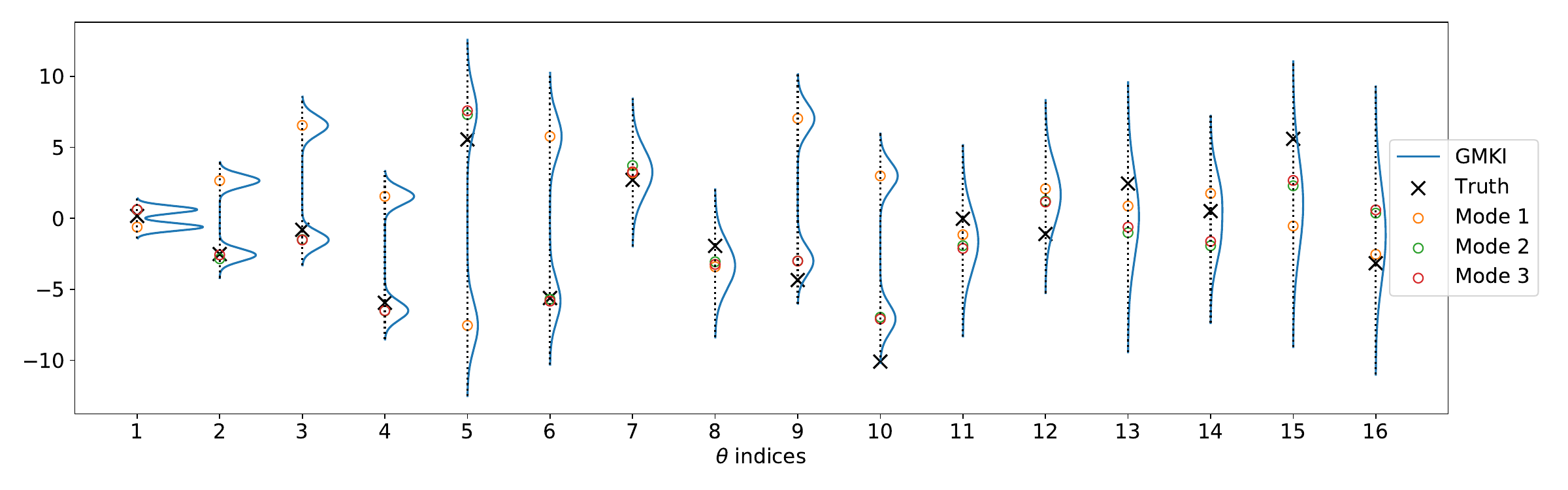}
    \caption{Navier-Stokes flow problem: the true Karhunen-Loeve expansion parameters $\theta_{(i)}$ (black crosses),  and mean estimations of $\theta_{(i)}$ for each modes~(circles) and the associated marginal distributions obtained GMKI at the $50$th iteration.}
    \label{fig:NS-2D-density}
\end{figure}

%%%%%%%%%%
We employ GMKI with $K=3$ modes, which are randomly initialized based on prior distribution with equal weights. 
Since we have $3$ modes and $N_\theta = 128$, in each iteration, we require $(2N_{\theta}+1 )K=771$ forward evaluations.
We depict the true initial vorticity field $\omega_0(x;\theta_{\rm ref})$, {its mirrored field~(the mirroring of the velocity field induces the antisymmetry in the vorticity field)} and the three recovered initial vorticity fields $\omega_0(x;m_k)$ obtained by GMKI at the $50$th iteration in~\cref{fig:NS-2D-vor}.  Mode 1 captures the mirroring field of $\omega_0(x;\theta_{\rm ref})$ and mode 2 and mode 3 capture $\omega_0(x;\theta_{\rm ref})$. \Cref{fig:NS-2D-convergence} presents the relative errors of the vorticity field, the optimization errors $\Phi_R(m_{n,k})$, the Frobenius norm $\lVert C_{n,k}\rVert_F$ and the Gaussian mixture weights $w_{n,k}$ (from left to right). It shows that our GMKI converges in fewer than 50 iterations. \Cref{fig:NS-2D-density} displays the
marginals of the estimated posterior distributions associated with the first $16$ theta coefficients obtained by GMKI. The marginal distributions exhibit clear bimodality, and the approximate posteriors have a high probability covering the true coefficients.

\section{Conclusion}
\label{sec:conclusion}
In this paper, we have presented a new framework for solving Bayesian inverse problems. The framework is based on the Fisher-Rao gradient flow.
Within this framework, we introduce a novel approach, Gaussian mixture Kalman inversion (GMKI), which leverages Gaussian mixtures and Kalman's methodology for numerical approximations of the the flow. GMKI is particularly useful when the posterior distribution has multiple modes and when the derivative of the forward model is not available or computationally expensive.

We derive the continuous time dynamics of the GMKI, showing its connection to Gaussian mixture variational inference, and studying its exploration effects and convergence properties.
Our numerical experiments showcase GMKI's capability in approximating posterior distributions with multiple modes. GMKI outperforms many existing Bayesian inference methods in terms of efficiency and accuracy. 

There can be numerous avenues for future research. On the algorithmic side, it is of interest to refine the approximations employed by GMKI in regions where the mixture components overlap significantly; see the experimental results in \Cref{ssec:1d-bimodal}. Moreover, although the Kalman methodology achieves a derivative-free implementation, it may suffer from degeneracy issues when the modes of the distribution concentrate on a low dimensional manifold; for a demonstration see \ref{appendix-GMKI-low-dim-manifold-posterior}. Therefore, improving the derivative-free methodology in such a scenario is important for enhancing GMKI.
On the theoretical
side, a thorough analysis of the convergence of GMKI for general target distribution could offer valuable insights for its practical application.

\section*{Acknowledgments} YC is supported by the Courant Instructorship. DZH is supported by the Fundamental Research Funds for the Central Universities and the high-performance computing platform of Peking
University. JH is supported by NSF grant DMS-2331096 and the Sloan research fellowship. DZH and AMS are supported by NSF award AGS1835860 and by the generosity of Eric and Wendy Schmidt by recommendation of the Schmidt Futures program.
AMS is supported by a Department of Defense Vannevar Bush 
Faculty Fellowship and by the SciAI Center, funded by the Office of Naval Research 
(ONR), under Grant Number N00014-23-1-2729.  SR is supported by Deutsche Forschungsgemeinschaft (DFG) - Project-ID 318763901 - SFB1294. We also thank the anonymous reviewers for their helpful suggestions.

\appendix
\section{Convergence of the Exploration-Exploitation Scheme for the Fisher-Rao Gradient Flow}
\label{append:convergence}

\begin{proof}[Proof of \Cref{prop:consistency}]
The solution of the Fisher-Rao gradient flow \cref{eq:mean-field-Fisher-Rao-intro} has an analytical solution
\begin{align}\label{e:rhot}
&\rho_t(\theta) =\frac{1}{Z_t} \rho_0(\theta)^{ e^{-t}}\rho_{\rm post}(\theta)^{1 - e^{-t}},\quad \frac{\rho_t(\theta)}{\rho_{\rm post}(\theta)}=\frac{1}{Z_t}\left(\frac{\rho_0(\theta)}{\rho_{\rm post}(\theta)}\right)^{e^{-t}}.
\end{align}
where $Z_t$ is the normalization constant.
The update \cref{eq:concept-alg} leads to
\begin{align}\label{e:rhot-discrete}
&\rho_n(\theta) = \frac{1}{Z_n} \rho_0(\theta)^{ (1 - \Delta t)^{n}}\rho_{\rm post}(\theta)^{1 - (1 - \Delta t)^{n}},\quad \frac{\rho_n(\theta)}{\rho_{\rm post}(\theta)}=\frac{1}{Z_n}\left(\frac{\rho_0(\theta)}{\rho_{\rm post}(\theta)}\right)^{(1 - \Delta t)^{n}},
\end{align}
where  $Z_n$ is the normalization constant. By comparing Eq.~\cref{e:rhot-discrete} and Eq.~\cref{e:rhot}, we have 
$$
\rho_n(\theta) = \rho_t(\theta) \quad \text{for}\quad t = -n\log(1 - \Delta t).
$$
Bringing $t = -n\log(1 - \Delta t)$ into \cref{e:KLconverge} leads to
\begin{align}
    \mathrm{KL}[\rho_{n} \Vert  \rho_{\rm post}] = \mathrm{KL}[\rho_{-n\log(1 - \Delta t)} \Vert  \rho]
    \leq (2+B + eB)K(1 - \Delta t)^n,
\end{align}
when $-n\log(1 - \Delta t) \geq \log((1 + B)K)$.
\end{proof}
% \section{Connections Between the Gaussian Kalman Inversion Approach and Gaussian Variational Inference}
% Stochastic linearization
% \label{append:connect GKI and VI}

\section{Gaussian Mixture Kalman Inversion Algorithm}
\label{sec:GMKI}
The detailed algorithm is presented in \cref{alg:GMKI}.
There are four hyperparameters: the number of mixtures $K$, the time-step size $\Delta t$, the number of particles $J$ for Monte Carlo integration, and the number of iterations $N$.
Increasing $K$ enhances the expressiveness of the Gaussian mixture model but also increases computational cost. Therefore, we recommend choosing $K$ based on available computational resources and prior knowledge of the number of target modes. Once all modes are captured, there is no need to increase $K$ further. A larger $J$ improves the accuracy of integration approximations, but it also increases computational cost. In the experiments, we adopt $J = 1000$.
The time step $\Delta t$ lies within the range $(0,1)$. While a larger $\Delta t$ accelerates convergence, it may also cause numerical instability. However, we found that $\Delta t = 0.5$ strikes a good balance between stability and efficiency. All of these parameters can be chosen adaptively, which we plan to explore in future work. Moreover, in our implementation, we store 
$\log w$ instead of $w$, and hence $w$ will never actually reach zero. Additionally, for efficiency, we avoid zero weights by setting a lower bound (e.g., $10^{-10}$) for $w$.

\begin{algorithm}
\caption{Gaussian Mixture Kalman Inversion}
\label{alg:GMKI}
\begin{algorithmic}
\State{Input: initial guess $\{w_{0,k}, m_{0,k}, C_{0,k}\}_{k=1}^{K}$, time-step size $\Delta t$, number of iterations $N$, number of particles $J$ for Monte Carlo, forward model $\F$}

\State{Output: final solution $\{w_{N,k}, m_{N,k}, C_{N,k}\}_{k=1}^{K}$}

\For{$n \gets 0$ to $N-1$}

%\State{// First step $\hat{\rho}_{n+1}(\theta) \propto \rho_n(\theta)^{1 - \Delta t}$}
\For{$k \gets 1$ to $K$}  
    \State{Sample $\{\theta^j\}_{j=1}^{J} \sim \N(\theta; m_{n,k}, \frac{C_{n,k}}{1 - \Delta t})$, first step update
 \begin{align*}
    &\hat{w}_{n+1,k} = \frac{1}{J}\sum_{j=1}^{J} f_{n,k}(\theta^j) 
    \\
    &\pmean_{n+1,k} = \frac{1}{\hat{w}_{n+1,k} J }\sum_{j=1}^{J} \theta^j f_{n,k}(\theta^j)
    \\
    &\pCov_{n+1,k} = \frac{1}{\hat{w}_{n+1,k}(J-1)}\sum_{j=1}^{J} (\theta^j - \pmean_{n+1,k})(\theta^j - \pmean_{n+1,k})^T f_{n,k}(\theta^j)
    \end{align*}
    }
    \EndFor

    \State{Normalize $\hat{w}_{n+1,k} := \frac{\hat{w}_{n+1,k}}{\sum_{k=1}^{K} \hat{w}_{n+1,k}}$}
   
   %\State{// Second step $\rho_{n+1}(\theta) \propto \hat{\rho}_{n+1}(\theta) e^{-\Delta t\Phi_R(\theta)}$}
    \For{$k \gets 1$ to $K$}    \Comment{Apply modified unscented transform \cite[Eq. 37]{huang2022efficient}}

    \State{Generate sigma-points $(a = \max\{\frac{1}{8}, \frac{1}{2N_{\theta}}\})$ }
    \begin{align*}
    &\theta_{k}^{0} = \pmean_{n+1, k} \\
    &\theta_{k}^j = \pmean_{n+1, k}  + \frac{1}{\sqrt{2a}} [\sqrt{\pCov_{n+1,k}}]_j \quad (1\leq j\leq N_\theta)\\
    &\theta_{k}^{j+N_\theta} = \pmean_{n+1, k} - \frac{1}{\sqrt{2a}} [\sqrt{\pCov_{n+1,k}}]_j\quad (1\leq j\leq N_\theta)
    \end{align*}
     \State{Approximate the mean and covariance}
    \begin{align*}
    &\bpy_{n+1,k} :=\F(\theta_{k}^{0}) \\
    &\pCov_{n+1,k}^{\theta x} := \sum_{j=1}^{2N_{\theta}} a(\theta_k^{j} - \pmean_{n+1, k})(\F(\theta_k^{j}) - \bpy_{n+1,k})^T \\
    &\pCov_{n+1,k}^{xx} := \sum_{j=1}^{2N_{\theta}} a(\F(\theta_k^{j}) - \bpy_{n+1,k})(\F(\theta_k^{j}) - \bpy_{n+1,k})^T +  \frac{1}{\Delta t} \Sigma_{\nu}
    \end{align*}
   \State{Second step update}
    \begin{equation*}
    \begin{split}
     m_{n+1,k} &= \pmean_{n+1,k} + \pCov_{n+1,k}^{\theta x} (\pCov_{n+1,k}^{x x})^{-1} (x - \bpy_{n+1,k})\\
     C_{n+1,k} &= \pCov_{n+1,k} - \pCov_{n+1,k}^{\theta x}(\pCov_{n+1,k}^{x x})^{-1} {\pCov_{n+1,k}^{{\theta x}^{T}}}\\
     w_{n+1,k} &=  \hat{w}_{n+1,k} e^{-\Delta t \Phi_R(\theta_k^{0})} 
    \end{split}
    \end{equation*}
    \EndFor

    \State{Normalize ${w}_{n+1,k} := \frac{{w}_{n+1,k}}{\sum_{k=1}^{K}  {w}_{n+1,k}}$}
   
\EndFor
\end{algorithmic}
\end{algorithm}

\section{Theoretical Studies About GMKI}
\label{append:GMKI}
\subsection{Proof of \cref{lem:GMKI-1-CTL}}
\label{proof:lem:GMKI-1-CTL}
Update equation \cref{eq:power-GMM} and the normalization of weights $\{\hat{w}_{n+1,k}\}_k$ can be combined and rewritten as 
\begin{subequations}
\label{eq:power-GMM-analysis}
\begin{align}
&\hat{w}_{n+1,k} =\frac{ w_{n,k}\int \N(\theta; m_{n,k}, C_{n,k}) \rho_n^{-\Delta t} \dd\theta}{\int \rho_n^{1-\Delta t} \dd\theta},\\
&\pmean_{n+1,k} = \frac{ \int \theta \N(\theta; m_{n,k}, C_{n,k}) \rho_n^{-\Delta t} \dd\theta}{\int \N(\theta; m_{n,k}, C_{n,k}) \rho_n^{-\Delta t} \dd\theta},\\
&\pCov_{n+1,k} = \frac{ \int (\theta - \pmean_{n+1,k})(\theta - \pmean_{n+1,k})^T \N(\theta; m_{n,k}, C_{n,k}) \rho_n^{-\Delta t} \dd\theta}{\int \N(\theta; m_{n,k}, C_{n,k}) \rho_n^{-\Delta t} \dd\theta}.
\end{align}
\end{subequations}
Now we derive the continuous time limit of the exploration step; we ignore the hat notation for simplicity. 
% The continuous time limit of $w_{t,k}$ ($w_{t,k}=w_{n\Delta t, k} = w_{n,k}$) is 
% \yc{(rewrite to make it rigorous; currently $\frac{{w}_{n+1,k} - w_{n,k}}{\Delta t}$ is not good; because $w_{n,k}$ also changes when $\Delta t$ changes. It is better to write in the way that ${w}_{n+1,k} - w_{n,k} = C\Delta t + O((\Delta t)^2)$; then according to the correspondance to numerical ODEs, we say its continuous limit is xxx)}
\begin{equation}
\begin{aligned}
 \frac{\hat{w}_{n+1,k} - w_{n,k}}{\Delta t} &= w_{n,k}\frac{ \int \Bigl[\N(\theta; m_{n,k}, C_{n,k}) - \rho_n \Bigr] \rho_n^{-\Delta t} \dd\theta}{\Delta t \int  \rho_n^{1-\Delta t} \dd\theta}
\\
&= w_{n,k}\frac{ \int \Bigl[\N(\theta; m_{n,k}, C_{n,k}) - \rho_n \Bigr] (1 - \Delta t\log\rho_n) \dd\theta}{\Delta t} + \bigO(\Delta t)
\\
&= -w_{n,k} \int \Bigl[\N(\theta; m_{n,k}, C_{n,k}) - \rho_n \Bigr] \log\rho_n \dd\theta + \bigO(\Delta t),
\end{aligned}
\end{equation}
where in the second identity, we used $\rho_n^{-\Delta t} = e^{-\Delta t\log \rho_n } = 1 - \Delta t\log \rho_n + \mathcal{O}(\Delta^2)$.
Therefore the continuous limit can be written as 
\[\dot{w}_{t,k}
    = -w_{t,k}\int \Bigl[\N(\theta; m_{t,k}, C_{t,k})  - \rho_t(\theta) \Bigr] \log \rho_t(\theta) \dd\theta \, .\]
Similarly, for $m_{n,k}$ we have
\begin{equation}
\label{C-mhat}
\begin{aligned}
 \frac{\pmean_{n+1,k} - m_{n,k}}{\Delta t}
&= \frac{ \int (\theta -  m_{n,k})\N(\theta; m_{n,k}, C_{n,k}) \rho_n^{-\Delta t} \dd\theta}{\Delta t\int \N(\theta; m_{n,k}, C_{n,k}) \rho_n^{-\Delta t} \dd\theta}
\\
&= \frac{ \int (\theta -  m_{n,k})\N(\theta; m_{n,k}, C_{n,k}) (1 - \Delta t\log\rho_n) \dd\theta}{\Delta t} + \bigO(\Delta t)
\\
&= -\int \N(\theta; m_{n,k}, C_{n,k}) (\theta - m_{n,k}) \log \rho_n(\theta) \dd\theta + \bigO(\Delta t)
\\
&= -C_{n,k}\int \N(\theta; m_{n,k}, C_{n,k}) \nabla_{\theta}\log \rho_n(\theta) \dd\theta + \bigO(\Delta t), 
\end{aligned}
\end{equation}
where in the last identity, we used integration by parts; it is also known as  the Stein's identity. Thus the continuous limit is 
\[\dot{m}_{t,k} = -\int \N(\theta; m_{t,k}, C_{t,k}) (\theta - m_{t,k}) \log \rho_t(\theta) \dd\theta .\]
Finally, for $C_{n,k}$, we have
\begin{equation}
\begin{aligned}
&\frac{\pCov_{n+1,k} - C_{n,k}}{\Delta t}
\\  
=&  \frac{ \int \Bigl[(\theta - \pmean_{n+1,k})(\theta - \pmean_{n+1,k})^T - C_{n,k}\Bigr] \N(\theta; m_{n,k}, C_{n,k}) \rho_n^{-\Delta t} \dd\theta}{\Delta t \int \N(\theta; m_{n,k}, C_{n,k}) \rho_n^{-\Delta t} \dd\theta}
\\
=&   \frac{ \int \Bigl[(\theta - \pmean_{n+1,k})(\theta - \pmean_{n+1,k})^T - C_{n,k}\Bigr] \N(\theta; m_{n,k}, C_{n,k}) (1 - \Delta t \log\rho_n) \dd\theta}
{
\Delta t \int \N(\theta; m_{n,k}, C_{n,k}) (1 - \Delta t \log\rho_n) \dd\theta
} + \bigO(\Delta t)
\\
=&   \frac{ \int \Bigl[(\theta - \pmean_{n+1,k})(\theta - \pmean_{n+1,k})^T - C_{n,k}\Bigr] \N(\theta; m_{n,k}, C_{n,k}) (1 - \Delta t \log\rho_n) \dd\theta}
{
\Delta t
} + \bigO(\Delta t)
\\
=&   -\int \Bigl[(\theta - m_{n,k})(\theta - m_{n,k})^T - C_{n,k}\Bigr] \N(\theta; m_{n,k}, C_{n,k})  \log\rho_n \dd\theta + \bigO(\Delta t)
\\
=& -C_{n,k}\int \N(\theta; m_{n,k}, C_{n,k}) \nabla_{\theta}\nabla_{\theta}\log \rho_n \dd\theta C_{n,k} + \bigO(\Delta t),
\end{aligned}
\end{equation}
where in the fourth identity, we used \cref{C-mhat} $\widehat{m}_{n+1,k} = m_{n,k} + \bigO(\Delta t)$. And in the last identity, we used integration by parts. Finally, this leads to the desired continuous limit stated in the proposition.

\subsection{Proof of \cref{lem:exploration-entropy}}
\label{proof:lem:exploration-entropy} 
The evolution equation of the entropy of $\rho_t$ is
\begin{equation}
    \begin{split}
        &\frac{\dd}{\dd t} \int -\rho_t \log \rho_t \dd\theta = -\int \frac{\dd\rho_t}{\dd t} \log\rho_t \dd\theta
        \\
        =& -\sum_{k} \dot{w}_{t,k} \int \N(\theta; m_{t,k}, C_{t,k})  \log\rho_t  \dd\theta
        \\
        &- \sum_{k}w_k \dot{m}_{t,k}^TC_{t,k}^{-1}\int  (\theta - m_{t,k}) \N(\theta; m_{t,k}, C_{t,k})  \log\rho_t  \dd\theta 
        \\
        &- \sum_{k}\frac{w_{t,k}}{2} {\rm tr}
        \Bigl[ 
        \dot{C}_{t,k}^T\int  \bigl( 
        C_{t,k}^{-1}(\theta - m_{t,k})(\theta - m_{t,k})^T C_{t,k}^{-1} - C_{t,k}^{-1}
        \bigr) \N(\theta; m_{t,k}, C_{t,k}) \log\rho_t \dd\theta 
        \Bigr]
        \\
        =& \sum_k \Bigl(\frac{\dot{w}_{t,k}^2}{ w_{t,k}} +  w_{t,k} \dot{m}_{t,k}^TC_{t,k}^{-1}\dot{m}_{t,k} + \frac{w_{t,k}}{2}{\rm tr}\Bigl[\dot{C}_{t,k} C_{t,k}^{-1}\dot{C}_{t,k} C_{t,k}^{-1}\Bigr]\Bigr).
    \end{split}
\end{equation}
Here we have used the fact $\int \frac{\dd\rho_t}{\dd t} \dd\theta= 0$ in the first identity. 
And in the second identity we used the fact that $\sum_{k} \dot{w}_{t,k} = 0$ and used the continuous time limit equations \cref{eq:GMKI-CTL-1}.

\subsection{Proof of \cref{lem:GMKI-CTL}}
\label{proof:lem:GMKI-CTL}
Combining \Cref{lem:GMKI-1-CTL} and \cref{eq:GMM-update-2} leads to the following  

\begin{equation}
\begin{split}
&\frac{m_{n+1,k} - m_{n,k}}{\Delta t} 
\\
=&
\frac{\pmean_{n+1,k} - m_{n,k}}{\Delta t}
+ \frac{\pCov_{n+1,k}^{\theta x} (\pCov_{n+1,k}^{x x})^{-1} (x - \bpy_{n+1,k}) }{\Delta t}
\\
=& -C_{n,k}\int \N(\theta; m_{n,k}, C_{n,k}) \nabla_{\theta}\log \rho_n(\theta) \dd\theta + \pCov_{n,k}^{\theta x} \Sigma_{\nu}^{-1} (x - \bpy_{n,k}) +\bigO(\Delta t). 
\\
 &\frac{C_{n+1,k} - C_{n,k}}{\Delta t} 
\\
=& \frac{\pCov_{n+1,k} - C_{n,k}}{\Delta t} - \frac{\pCov_{n+1,k}^{\theta x}(\pCov_{n+1,k}^{x x})^{-1} {\pCov_{n+1,k}^{{\theta x}^{T}}}}{\Delta t}  
\\
=&  -C_{n,k}\int \N(\theta; m_{n,k}, C_{n,k}) \nabla_{\theta}\nabla_{\theta}\log \rho_n(\theta) \dd\theta C_{n,k} - \pCov_{n,k}^{\theta x}\Sigma_{\nu}^{-1} {\pCov_{n,k}^{{\theta x}^{T}}} + \bigO(\Delta t). 
\end{split}
\end{equation}
Using the above formula, we obtain the continuous limit. For the weights, we have
\begin{subequations}
\begin{align}
&w_{n+1,k} =\frac{ \hat{w}_{n+1,k} \int \N(\theta; \pmean_{n+1,k}, \pCov_{n+1,k}) e^{-\Delta t \Phi_R(\theta)} \dd\theta}{\int \hat{\rho}_{n+1}e^{-\Delta t \Phi_R(\theta)} \dd\theta}.
\end{align}
\end{subequations}
Therefore,
\begin{equation}
\begin{split}
&\frac{w_{n+1,k} - w_{n,k}}{\Delta t} = 
 \frac{w_{n+1,k} - \hat{w}_{n+1,k}}{\Delta t } + \frac{\hat{w}_{n+1,k} - w_{n,k}}{\Delta t}
\\
=& \hat{w}_{n+1,k}\frac{ \int \bigl[\N(\theta; \pmean_{n+1,k}, \pCov_{n+1,k})  - \hat{\rho}_{n+1}\bigr]e^{-\Delta t \Phi_R}  \dd\theta}{\Delta t\int \hat{\rho}_{n+1}e^{-\Delta t \Phi_R} \dd\theta} +\frac{\hat{w}_{n+1,k} - w_{n,k}}{\Delta t} + \bigO(\Delta t)
\\
=&\hat{w}_{n+1,k}\frac{ \int \bigl[\N(\theta; \pmean_{n+1,k}, \pCov_{n+1,k})  - \hat{\rho}_{n+1}\bigr](1 - \Delta t \Phi_R)  \dd\theta}{\Delta t} + \frac{\hat{w}_{n+1,k} - w_{n,k}}{\Delta t} + \bigO(\Delta t)
\\
=& -w_{n,k}\int \Bigl[\N(\theta; m_{n,k}, C_{n,k}) - \rho_n \Bigr] (\log\rho_n  + \Phi_R) \dd\theta + \bigO(\Delta t),
\end{split}
\end{equation}
from which we readily obtain the continuous limit for the equation of the weights. Here in the last step we used the result in \ref{proof:lem:GMKI-1-CTL}.

\subsection{Proof of \cref{lem:GMKI-linear}}
\label{proof:lem:GMKI-linear} 
Under the Gaussian posterior assumption (i.e. $\Phi_R$ is quadratic) and based on the formula in \eqref{eqn:stein-identity1} and \eqref{eqn:stein-identity2}, we get
\begin{align}
\label{eq:Kalman_Gaussian_posterior}
    &\pCov_{t, k}^{\theta x} \Sigma_{\nu}^{-1} (x - \bpy_{t,k})= -C_{t,k}\int  \N(\theta; m_{t,k}, C_{t,k}) \nabla_{\theta} \Phi_R \dd\theta,
    \\
    &\pCov_{t, k}^{\theta x}\Sigma_{\nu}^{-1} {\pCov_{t,k}^{T}}= C_{t,k} \left(\int \N(\theta; m_{t,k}, C_{t,k}) \nabla_{\theta} \nabla_{\theta} \Phi_R \dd\theta \right) C_{t,k}^T.
\end{align}
Bringing \cref{eq:Kalman_Gaussian_posterior} into the continuous time dynamics~\cref{eq:GMKI-CTL} and using integration by parts (also known as Stein's identities) leads to 
\begin{subequations}
\label{appendix: continuous time eq}
\begin{align}
    \dot{m}_{t,k} =& -C_{t,k}\int \N(\theta; m_{t,k}, C_{t,k})\nabla_{\theta} \bigl( \log \rho_t + \Phi_R\bigr)\dd\theta  \nonumber\\
    =& -\int \N(\theta; m_{t,k}, C_{t,k})(\theta - m_{t,k}) \bigl( \log \rho_t + \Phi_R \bigr)\dd\theta, 
    \label{appendix: continuous time eq mean}\\
    \dot{C}_{t,k} =& -\int \N(\theta; m_{t,k}, C_{t,k})\nabla_{\theta}\nabla_{\theta} \bigl(\log\rho_t + \Phi_R\bigr) \dd\theta  \nonumber\\
    =& -\int \N(\theta; m_{t,k}, C_{t,k})\Bigl((\theta - m_{t,k})(\theta - m_{t,k})^{T}  - C_{t,k}\Bigr) \bigl(\log\rho_t + \Phi_R\bigr) \dd\theta,  
    \label{appendix: continuous time eq cov}\\
    \dot{w}_{t,k} =&-w_{t,k} \int \Bigl[ \N(\theta; m_{t,k}, C_{t,k})  - \rho_t \Bigr] \Bigl[ \log \rho_t + \Phi_R \Bigr]\dd\theta. 
\end{align}
\end{subequations}
The evolution equation of the KL divergence between $\rho_t$ and $\rho_{\rm post}$ is then
\begin{equation}
    \begin{split}
        &\frac{\dd }{\dd t}\mathrm{KL}[\rho_t \Vert \rho_{\rm post}] \\
        =& \frac{\dd}{\dd t}\int \rho_t\log\Bigl(\frac{\rho_t}{\rho_{\rm post}}\Bigr) \dd\theta= \int \frac{\dd\rho_t}{\dd t}\bigl(\log\rho_t + \Phi_R\bigr) \dd\theta
        \\
        =& \sum_{k} \dot{w}_{t,k} \int \N(\theta; m_{t,k}, C_{t,k}) \bigl(\log\rho_t + \Phi_R\bigr) \dd\theta
        \\
        &+ \sum_{k}w_k \dot{m}_{t,k}^TC_{t,k}^{-1}\int  (\theta - m_{t,k}) \N(\theta; m_{t,k}, C_{t,k}) \bigl(\log\rho_t + \Phi_R\bigr) \dd\theta 
        \\
        &+ \sum_{k}\frac{w_{t,k}}{2} {\rm tr}
        \Bigl[ 
        \dot{C}_{t,k}^T\int  \bigl( 
        C_{t,k}^{-1}(\theta - m_{t,k})(\theta - m_{t,k})^T C_{t,k}^{-1} - C_{t,k}^{-1}
        \bigr) \N(\theta; m_{t,k}, C_{t,k}) \bigl(\log\rho_t + \Phi_R\bigr) \dd\theta 
        \Bigr]
        \\
        =& -\sum_k \Bigl(\frac{\dot{w}_{t,k}^2}{ w_{t,k}} +  w_{t,k} \dot{m}_{t,k}^TC_{t,k}^{-1}\dot{m}_{t,k} + \frac{w_{t,k}}{2}{\rm tr}\Bigl[\dot{C}_{t,k} C_{t,k}^{-1}\dot{C}_{t,k} C_{t,k}^{-1}\Bigr]\Bigr).
    \end{split}
\end{equation}
Here in the last identity, we used the equation of the continuous time dynamics in \eqref{appendix: continuous time eq} to simplify the formula.

Consider any stationary point $\rho_{\infty} = \sum_k w_{k,\infty}\N(\theta; m_{k,\infty}, C_{k,\infty})$ with nonzero $w_{k,\infty}$. 
% \yc{(please add more explanations; you are using \eqref{appendix: continuous time eq} right? there are three equations; here you use two.)}
The stationary point condition for the mean $m_{k,\infty}$ \eqref{appendix: continuous time eq mean} is 
\begin{equation}
\label{eq:stationary_condition-mean}
\int \N(\theta; m_{k,\infty}, C_{k,\infty})\nabla_{\theta} \log \rho_{\infty} \dd\theta = -C_{\rm post}^{-1}(m_{k,\infty} - m_{\rm post}), 
\end{equation}
where we used that $\int \N(\theta; m_{\infty,k}, C_{\infty,k}) \nabla_{\theta}\Phi_R\dd\theta  = C_{\rm post}^{-1}(m_{k,\infty} - m_{\rm post})$.
The stationary point condition for the covariance $C_{k,\infty}$ \eqref{appendix: continuous time eq cov} is 
\begin{equation}
\label{eq:stationary_condition-cov}
\int \N(\theta; m_{k,\infty}, C_{k,\infty})\nabla_{\theta}\nabla_{\theta} \log \rho_{\infty} \dd\theta = -C_{\rm post}^{-1},
\end{equation}
where we used that $\int \N(\theta; m_{\infty,k}, C_{\infty,k}) \nabla_{\theta}\nabla_{\theta}\Phi_R\dd\theta  =  C_{\rm post}^{-1}$.
Multiplying  \cref{eq:stationary_condition-mean} and \cref{eq:stationary_condition-cov}  by $w_{k,\infty}$ and summing the results yields
    \begin{equation}
    \begin{aligned}
        &m_{\rm post} - \sum_{k}w_{k,\infty}m_{k,\infty}   
        = C_{\rm post}\int\sum_{k}w_{k,\infty}\N(\theta; m_{k,\infty}, C_{k,\infty})\nabla_{\theta}\log \rho_{\infty}  \dd\theta  = 0, \\
        &C_{\rm post}^{-1} = -\int \sum w_{k,\infty}\N(\theta; m_{k,\infty}, C_{k,\infty})\nabla_{\theta}\nabla_{\theta} \log \rho_{\infty} \dd\theta = {\rm FIM}[\rho_{\infty}].
    \end{aligned}
    \end{equation}

\subsection{Derivation of the Simplified Continuous Time Dynamics~\cref{eq:GMKI-CTL-GM}}
\label{append:GMKI-simplified}
% \yc{(Need to rewrite. Seems too may approximation steps in the writing. Can be made more clear.)}
{In this section, we formally derive \cref{eq:GMKI-CTL-GM}, assuming these Gaussian components in $\rho_{\rm post}$ are well separated.}
When $\theta$ is close to  $m_k^{*}$, one may make the following approximation 
\begin{align}
\label{eq:GM-posterior-disjoint}
&\Phi_R(\theta)  = \log \rho_{\rm post}(\theta) \approx -\log \N(\theta, m_k^{*}, C_k^{*}) - \log(w_k^{*}).
\end{align}
Combining the definition of $\Phi_R$ in~\cref{eq:Phi_R_F} with \cref{eq:GM-posterior-disjoint} leads to that
\begin{equation}
\label{eq:GM-posterior-disjoint-eq}
    -\frac{1}{2}(x - \F(\theta))^T\Sigma_{\nu}^{-1}(x - \F(\theta)) \approx -\frac{1}{2}(\theta - m_k^{*})^TC_{k}^{*^{-1}}(\theta - m_k^{*}) + \textrm{constant}.
\end{equation} 
This implies that $\F(\theta)$ is approximately locally linear around $m=m_k^*$ with $\F(\theta) \approx F_k \theta + c$, such that $F_k^T\Sigma_{\nu}^{-1}F_k = C_k^{*^{-1}}$ and $C_k^{*} F_k^{T}\Sigma_{\nu}^{-1}(x - c) = m_k^{*}$.
{Based on the above derivation, when the $k$-the component $\N(\theta; m_{t,k},C_{t,k})$ in the Gaussian mixture approximation is concentrating around $m_k^{*}$,} the expectation and covariance in the continuous time limit of the proposed GMKI~\eqref{eq:E_cov_F} can be approximated as 
\begin{subequations}
\label{eq:hat_x_C}
\begin{align}
    \bpy_{t,k} &= \E[\F(\theta)] \approx \E[F_k \theta + c] = F_k m_{t,k} + c,
\\
\pCov_{t, k}^{\theta x}&=\Cov \bigl[ \theta,\F(\theta)\bigr]
\approx\E\bigl[\bigl(\theta-m_{t,k}\bigr)\otimes\bigl(F_k(\theta - m_{t,k})\bigr)\bigr] = C_{t,k}F_k^T.
\end{align}
\end{subequations}
{Here the expectation are taken with respect to $\N(\theta; m_{t,k},C_{t,k})$.}

Now we will simplify \cref{eq:GMKI-CTL} by neglecting the interaction between well separated Gaussian components to obtain \cref{eq:GMKI-CTL-GM}.
For the mean evolution equation~\eqref{eq:GMKI-CTL-m}, we have 
\begin{equation}
\label{eq:GMKI-CTL-m-reduce}
    \begin{split}
    \dot{m}_{t,k} 
    \approx& -C_{t,k}\int \N(\theta; m_{t,k}, C_{t,k}) \nabla_{\theta} \log \bigl(w_k \N(\theta; m_{t,k}, C_{t,k})\bigr)\dd\theta + \pCov_{t, k}^{\theta x} \Sigma_{\nu}^{-1} (x - \bpy_{t,k})
    \\
    =& \pCov_{t, k}^{\theta x} \Sigma_{\nu}^{-1} (x - \bpy_{t,k})
    \\
    \approx& C_{t,k} (C_k^{*})^{-1}(m_k^{*} - m_{t,k}).  
    \end{split}
\end{equation}
The first approximation is obtained by substituting $\log\rho_t$ with $\log (w_k \N(\theta; m_{t,k}, C_{t,k}))$, due to the well separateness assumption; the resulting integral is zero so leads to the second identity. The third approximation is obtained by using \cref{eq:hat_x_C} and the relation $F_k^T\Sigma_{\nu}^{-1}F_k = C_k^{*^{-1}}$ and $C_k^{*} F_k^{T}\Sigma_{\nu}^{-1}(x - c) = m_k^{*}$. 

For the covariance evolution equation~\eqref{eq:GMKI-CTL-C}, similarly we have 
\begin{equation}
\label{eq:GMKI-CTL-C-reduce}
\begin{split}
    \dot{C}_{t,k} \approx& -C_{t,k}\left(\int \N(\theta; m_{t,k}, C_{t,k}) \nabla_{\theta}\nabla_{\theta}\log\bigl(w_k \N(\theta; m_{t,k}, C_{t,k})\bigr)  \dd\theta\right) C_{t,k} - \pCov_{t, k}^{\theta x}\Sigma_{\nu}^{-1} {\pCov_{t,k}^{T}} 
    \\
    =&\  C_{t,k} - \pCov_{t, k}^{\theta x}\Sigma_{\nu}^{-1} {\pCov_{t,k}^{T}}
    \\
     =&\ C_{t,k} - C_{t,k} (C_k^{*})^{-1} C_{t,k}.  
\end{split}
 \end{equation}
 The first approximation is obtained by substituting $\log\rho_t$ with $\log (w_k \N(\theta; m_{t,k}, C_{t,k}))$. The third approximation is obtained by using \cref{eq:hat_x_C}. 
 
Finally, for the weight evolution equation~\eqref{eq:GMKI-CTL-w}, using the formula $\rho_t(\theta) = \sum_{i}\int w_{t,i} \N(\theta; m_{t,i}, C_{t,i})$, we get
\begin{equation}
\label{eq:GMKI-CTL-w-reduce}
\begin{split}
    \dot{w}_{t,k} 
    =& -w_{t,k} \int \N(\theta; m_{t,k}, C_{t,k}) \log \rho_t(\theta) \dd\theta  
     + w_{t,k} \sum_{i}\int w_{t,i} \N(\theta; m_{t,i}, C_{t,i}) \log \rho_t(\theta) \dd\theta  
     \\
    &-w_{t,k} \int \N(\theta; m_{t,k}, C_{t,k})\Phi_R(\theta) \dd\theta 
    + w_{t,k} \sum_{i} \int w_{t,i} \N(\theta; m_{t,i}, C_{t,i})  \Phi_R(\theta) \dd\theta 
    \\
    \approx& -w_{t,k} \int \N(\theta; m_{t,k}, C_{t,k}) \log [w_{t,k} \N(\theta; m_{t,k}, C_{t,k})] \dd\theta  
    \\&+ w_{t,k} \sum_{i}\int w_{t,i} \N(\theta; m_{t,i}, C_{t,i}) \log [w_{t,i} \N(\theta; m_{t,i}, C_{t,i})] \dd\theta \\
     &+w_{t,k} \int \N(\theta; m_{t,k}, C_{t,k})\Bigl(\log \N(\theta, m_k^{*}, C_k^{*}) + \log(w_k^{*})\Bigr) \dd\theta 
     \\
    &-w_{t,k} \sum_i \int w_{t,i} \N(\theta; m_{t,i}, C_{t,i}) \Bigl(\log \N(\theta, m_i^{*}, C_i^{*}) + \log(w_i^{*})\Bigr) \dd\theta 
     \\
     =& w_{t,k}\bigl(\log w_{k}^{*} - \log w_{t,k} - \sum_{i} w_{t,i}(\log w_{i}^{*} - \log w_{t,i})\bigr).   
\end{split}
\end{equation}
The first approximation is obtained by substituting $\log\rho_t$ with $\log (w_k \N(\theta; m_{t,k}, C_{t,k}))$ and using \cref{eq:GM-posterior-disjoint} for approximating $\Phi_R(\theta)$. {Combining \cref{eq:GMKI-CTL-m-reduce,eq:GMKI-CTL-C-reduce,eq:GMKI-CTL-w-reduce} leads to the simplified continuous time dynamics~\cref{eq:GMKI-CTL-GM}.
}

\subsection{Proof of \cref{lem:GMKI-simplified}}
\label{append:proof:GMKI-simplified}
The mean, covariance and weight evolution equations~\cref{eq:GMKI-CTL-GM-m,eq:GMKI-CTL-GM-C,eq:GMKI-CTL-GM-w} admit analytical solutions 

\begin{subequations}
\begin{align}
   &m_{t,k} = m_k^{*} + e^{-t}\Bigl((1-e^{-t})C_{k}^{*^{-1}} + e^{-t}C_k(0)^{-1}\Bigr)^{-1} C_k(0)^{-1} \bigl(m_k(0) - m_k^{*}\bigr),   
\\
&C_{t,k}^{-1} =  C_{k}^{*^{-1}} + e^{-t}(C_k(0)^{-1} - C_{k}^{*^{-1}} ), \\
&w_k 
     =\frac{w_k^{*} \Bigl(\frac{w_k(0)}{w_k^{*}} \Bigr)^{e^{-t}}}
{ \sum_i w_i^{*} \Bigl(\frac{w_i(0)}{w_i^{*}} \Bigr)^{e^{-t}}  }. 
\end{align}
\end{subequations} 
They will converge to $m_k^{*}$, $C_k^{*}$ and $w_k^{*}$ exponentially fast.

\subsection{Proof of \Cref{lem:GMKI-affine-invariance}}
\label{append:proof:GMKI-affine-invariance}
   Consider any invertible affine mapping $\varphi : \theta \rightarrow \widetilde\theta = A\theta + b$, and define corresponding vector and matrix transformations
\begin{align*}
&\widetilde{m}_{t,k} = Am_{t,k} + b \quad \widetilde{C}_{t,k} = AC_{t,k} A^T,
\end{align*}
density transformations
\begin{align*}
{\widetilde{\rho}}(\widetilde\theta) = \varphi_{\sharp} \rho(\theta)  =  \rho(A^{-1}(\widetilde\theta - b))|A|^{-1}
\quad \N(\widetilde\theta; \widetilde m_{t,k}, \widetilde C_{t,k} ) =  \N( \theta;   m_{t,k},  C_{t,k} )|A|^{-1},
\end{align*}
function transformations
\begin{align*}
\widetilde{\F}(\widetilde\theta) =  \F(A^{-1}(\widetilde\theta - b)) \quad \quad \widetilde{\Phi}_R(\widetilde\theta) =  \Phi_R(A^{-1}(\widetilde\theta - b)),
\end{align*}
and their related expectation and covariance
\begin{align*}
&\widetilde{x}_{t, k} =\E[\widetilde\F(\widetilde\theta)],
\qquad 
\widetilde{C}^{\theta x}_{t, k} = \mathrm{Cov}[\widetilde\theta, \widetilde\F(\widetilde\theta)], \quad \textrm{with} \quad \widetilde\theta \sim \N(\widetilde{m}_{t,k}, \widetilde{C}_{t,k}),
\end{align*}
then we have 
\begin{equation}
\begin{split}
&\nabla_{\widetilde{\theta}}\log\widetilde{\rho}(\widetilde{\theta}) = A^{-T}\nabla_{\theta}\log \rho(\theta) 
\qquad
\nabla_{\widetilde{\theta}}\nabla_{\widetilde{\theta}}\log\widetilde{\rho}(\widetilde{\theta}) = A^{-T}\nabla_{\theta}\log \rho(\theta) A^{-1}
\\
&\widetilde{x}_{t, k} =\bpy_{t,k}
\qquad 
\widetilde{C}_{t, k}^{\theta x} = A\pCov_{t, k}^{\theta x}.
\end{split}
\end{equation}
The evolution equations of $\widetilde{m}_{t,k}$, $\widetilde{C}_{t,k}$, $w_{t,k}$ in \cref{eq:GMKI-CTL} can be rewritten as
\begin{align*}
    \dot{\widetilde{m}}_{t,k} =& -AC_{t,k}\int \N(\theta; m_{t,k}, C_{t,k}) \nabla_{\theta} \log \rho_t(\theta) \dd\theta + A\pCov_{t, k}^{\theta x} \Sigma_{\nu}^{-1} (x - \bpy_{t,k}), \\
    =& -\widetilde{C}_{t,k}\int \N(\widetilde\theta; \widetilde{m}_{t,k}, \widetilde{C}_{t,k}) \nabla_{\widetilde\theta} \log \widetilde\rho_t(\widetilde\theta) \dd\widetilde\theta + \widetilde{C}_{t,k}^{\theta x}\Sigma_{\nu}^{-1} (x - \widetilde{x}_{t,k}), 
    \\
    \dot{\widetilde{C}}_{t,k} =& -AC_{t,k}\left(\int \N(\theta; m_{t,k}, C_{t,k}) \nabla_{\theta}\nabla_{\theta}\log \rho_t(\theta)  \dd\theta\right) C_{t,k}A^T - A\pCov_{t, k}^{\theta x}\Sigma_{\nu}^{-1} {\pCov_{t,k}^{{\theta x}T}}A^T,\\
    =& -\widetilde{C}_{t,k}\left(\int \N(\widetilde{\theta}; \widetilde{m}_{t,k}, \widetilde{C}_{t,k}) \nabla_{\widetilde\theta}\nabla_{\widetilde \theta}\log \widetilde\rho_t(\widetilde{\theta})  \dd\widetilde{\theta}\right) \widetilde{C}_{t,k} - \widetilde{C}_{t, k}^{\theta x}\Sigma_{\nu}^{-1} {\widetilde{C}_{t,k}^{{\theta x}^T}},
    \\
    \dot{w}_{t,k} =&-w_{t,k} \int \Bigl[ \N(\theta; m_{t,k}, C_{t,k})  - \rho_t(\theta) \Bigr] \Bigl[ \log \rho_t(\theta) + \Phi_R(\theta) \Bigr]\dd\theta\\
    =&-w_{t,k} \int \Bigl[ \N(\widetilde\theta; \widetilde{m}_{t,k}, \widetilde{C}_{t,k})  - \widetilde\rho_t(\widetilde{\theta}) \Bigr] \Bigl[ \log \widetilde\rho_t(\widetilde{\theta}) + \widetilde{\Phi}_R(\widetilde{\theta}) \Bigr]\dd\widetilde{\theta}.
\end{align*}
Hence the continuous time limit equation \cref{eq:GMKI-CTL} of GMKI is affine invariant.

\subsection{Connections Between the GMKI Approach and Gaussian Mixture Variational Inference}
\label{appendix:connect GMKI and GMVI}
Gaussian mixture variational inference seeks to identify a minimizer of ${\rm KL}[\rho^{\rm GM} \Vert \rho_{\rm post}]$, where \[\rho^{\rm GM}(\theta; a) = \sum_{k=1}^{K} w_k \N(\theta; m_k, C_k)\] is a $K$-component Gaussian mixture, parameterized by their means, covariances and weights denoted by \[a:=[m_1, ..., m_k, ..., m_K, C_1. ..., C_k, ..., C_K, w_1, ..., w_k, ..., w_K].\] 
The derivatives of the KL divergence with respect to $a$ are
\begin{subequations}
\label{eq:partial_KL_partial_a}
\begin{align}
&\frac{\partial {\rm KL}[\rho^{\rm GM}(\cdot; a) \Vert \rho_{\rm post}]}{\partial m_k} = w_k\int \N(\theta; m_k, C_k) \Bigl( \nabla_{\theta} \log\rho^{\rm GM}  -  \nabla_{\theta} \log\rho_{\rm post} \Bigr)  \dd\theta,
\\
& \frac{\partial{\rm KL}[\rho^{\rm GM}(\cdot; a) \Vert \rho_{\rm post}]}{\partial C_k} = \frac{w_k}{2}\int \N(\theta; m_k, C_k) \Bigl(\nabla_{\theta}\nabla_{\theta}\log \rho^{\rm GM}  - \nabla_{\theta}\nabla_{\theta}\log \rho_{\rm post}\Bigr) \dd\theta,
\\
&\frac{\partial{\rm KL}[\rho^{\rm GM}(\cdot; a) \Vert \rho_{\rm post}]}{\partial w_k} = \int \N(\theta; m_k, C_k) \Bigl(\log\frac{\rho^{\rm GM}}{\rho_{\rm post}} + 1\Bigr) \dd\theta.
\end{align}
\end{subequations}

The algorithm of natural gradient descent uses the finite dimensional version of the Fisher-Rao metric tensor in the parameter space, also known as the Fisher information matrix with the form \begin{equation}
    {\rm FI}(a) = \int \frac{\nabla_a \rho^{\rm GM}(\theta; a)  \otimes \nabla_a \rho^{\rm GM}(\theta; a) }{\rho^{\rm GM}(\theta; a) }\dd\theta
\end{equation} 
as the preconditioner for gradient descent. Here, we write down its continuous limit, namely the natural gradient flow. To do so, we use the perspective of proximal point method and consider
\begin{subequations}
\begin{align}
    a_{n+1} = &\argmin_{a} {\rm KL}[\rho^{\rm GM}(\cdot; a) \Vert \rho_{\rm post}] + \frac{1}{2\Delta t}\bigl\langle a-a_n, {\rm FI}(a_n) (a - a_n) \bigr\rangle,
    \\
    &\mathrm{\ \ subject\ to\  }\sum_{k=1}^{K} w_{n+1, k} = 1.
\end{align}
\end{subequations}
{By using the formula of derivatives in \cref{eq:partial_KL_partial_a}, the Lagrangian multiplier to handle the constraint in the above optimization, and taking $\Delta t \rightarrow 0$, we arrive at the following natural gradient flow}
\begin{align}
\label{eq:FR-GM}
  \begin{bmatrix}
  \dot{m}_{k} \\
  \dot{C}_{k} \\
  \dot{w}_{k} 
  \end{bmatrix}
  =({\rm FI}(a))^{-1}
  \begin{bmatrix}
  -w_k\int \N(\theta; m_k, C_k) \Bigl( \nabla_{\theta} \log\rho^{\rm GM}  -  \nabla_{\theta} \log\rho_{\rm post} \Bigr)  \dd\theta
  \\
  -\frac{w_k}{2}\int \N(\theta; m_k, C_k) \Bigl(\nabla_{\theta}\nabla_{\theta}\log \rho^{\rm GM}  - \nabla_{\theta}\nabla_{\theta}\log \rho_{\rm post}\Bigr) \dd\theta
  \\
  -\int \bigl(\N(\theta; m_k, C_k) - \rho^{\rm GM}\bigr) \log\bigl(\frac{\rho^{\rm GM}}{\rho_{\rm post}}\bigr) \dd\theta 
  \end{bmatrix}.
\end{align}
Computation of ${\rm FI}(a)$ is costly.
For better efficiency, diagonal approximations of the Fisher information matrix have been used in the literature~\cite{lin2019fast}, which leads to 
\begin{equation}
\label{eq:FI}
{\rm FI}(a) \approx \textrm{diag}\left(w_1 C_1^{-1}, ..., w_k C_k^{-1}, ..., w_K C_K^{-1}, w_1X_1, ..., w_kX_k, ..., w_KX_K, \frac{1}{w_1}, ..., \frac{1}{w_k}, ..., \frac{1}{w_K}\right).
\end{equation}
where each $X_k$ is a 4-th order tensor satisfying
\begin{equation}
\label{eq:FI-X}
X_k Y =     \frac{1}{4}C_k^{-1} (Y + Y^T) C_k^{-1}, \quad \forall \ Y \in \R^{N_{\theta}\times N_{\theta}}.  
\end{equation}
Bringing the approximated Fisher information matrix \cref{eq:FI} into the natural gradient flow \cref{eq:FR-GM} leads to the following equation:
\begin{equation}
\begin{split}
    \label{eq:Appr-FR-GM}
        \dot{m}_{k} 
        &= -C_k\int \N(\theta; m_k, C_k) \Bigl( \nabla_{\theta} \log\rho^{\rm GM}  -  \nabla_{\theta} \log\rho_{\rm post} \Bigr)  \dd\theta,
        \\
        \dot{C}_{k} 
        &= -C_k\Bigl(\int \N(\theta; m_k, C_k)\bigl(\nabla_{\theta}\nabla_{\theta}\log \rho^{\rm GM}  - \nabla_{\theta}\nabla_{\theta}\log \rho_{\rm post}\bigr) \dd\theta\Bigr) C_k,
        \\
        \dot{w}_{k} &= -w_k\int \Bigl(\N(\theta; m_k, C_k) -  \rho^{\rm GM}\Bigr)\log\bigl(\frac{\rho^{\rm GM}}{\rho_{\rm post}}\bigr) \dd\theta. 
\end{split}
\end{equation}
This is the natural gradient flow with diagonal approximations of the Fisher information matrix; its discretization is the approximate natural gradient descent algorithm that has been used in the literature.

By comparing \cref{eq:Appr-FR-GM} with the continuous-time limit of our GMKI as presented in \cref{eq:GMKI-CTL}, we observe that our GMKI can be seen as a derivative-free approximation of the approximate natural gradient descent. The approximation is made through stochastic linearization, for $\nabla_{\theta}\log \rho_{\rm post}$ and $\nabla_{\theta}\nabla_{\theta}\log \rho_{\rm post}$, based on the Kalman methodology explained in \cref{remark-connecting Kalman and VI}.

{For completeness, in the following part, we present a derivation of the diagonal approximations of the Fisher information matrix~\eqref{eq:FI}.}
Let $\N_k$ denote $\N(\theta; m_k, C_k)$ and {$\delta_{k,i}$ be the indicator function which is zero if and only if $k=i$}. We can get \cref{eq:FI} by only keeping the diagonal blocks of ${\rm FI}(a)$ and approximating the diagonals under the assumptions that different Gaussian components are well separated. More precisely, for the diagonal block regarding the weight $\{w_k\}$, we have 
\begin{equation}
\int \frac{\nabla_{w_k} \rho^{\rm GM} \otimes \nabla_{w_i} \rho^{\rm GM} }{\rho^{\rm GM} }\dd\theta
     = \int \frac{\N_k \N_i}{\rho^{\rm GM} }\dd\theta \approx \delta_{k,i}\int \frac{\N_k \N_k  }{w_k \N_k}\dd\theta = \frac{\delta_{k,i}}{w_k}.
\end{equation}
{Here we substitute $\rho^{\rm GM}$ by $w_k \N_k$ during its integration with $\N_k$. We note that we will keep using this approximation multiple times in the following derivations.}

For the diagonal block regarding the mean $m_k$, we have
\begin{align}
     \int \frac{\nabla_{m_k} \rho^{\rm GM} \otimes \nabla_{m_i} \rho^{\rm GM} }{\rho^{\rm GM} }\dd\theta =& \int \frac{ w_k w_i \N_k\N_i C_k^{-1}(\theta - m_k)(\theta - m_i)^T C_i^{-1} }{\rho^{\rm GM} }\dd\theta
     \\
     \approx& \delta_{k,i}\int \frac{ w_k^2  \N_k^2 C_k^{-1}(\theta - m_k)(\theta - m_k)^T C_k^{-1} }{w_k\N_k }\dd\theta \nonumber
     \\
     =& \delta_{k,i} w_k C_k^{-1}. \nonumber
\end{align}

For the diagonal block regarding the covariance ${C_k}$, we have
\begin{equation}
\label{eq:FI-C-approximation}
\begin{split}
     \int &\frac{\nabla_{C_k} \rho^{\rm GM} \otimes \nabla_{C_i} \rho^{\rm GM} }{\rho^{\rm GM} }\dd\theta 
     \\
     =& \int \frac{w_kw_i\N_k\N_i \Bigl( C_k^{-1}(\theta - m_k)(\theta - m_k)^TC_k^{-1} - C_k^{-1} \Bigr)\otimes\Bigl( C_i^{-1}(\theta - m_i)(\theta - m_i)^TC_i^{-1} - C_i^{-1} \Bigr)}{4\rho^{\rm GM}} \dd\theta
     \\
     \approx& \delta_{k,i}\int \frac{w_k^2\N_k^2 \Bigl( C_k^{-1}(\theta - m_k)(\theta - m_k)^TC_k^{-1} - C_k^{-1} \Bigr)\otimes\Bigl( C_k^{-1}(\theta - m_k)(\theta - m_k)^TC_k^{-1} - C_k^{-1} \Bigr)}{4w_k\N_k} \dd\theta 
     \\
     =& \frac{\delta_{k,i}}{4}\int w_k\N_k \Bigl( C_k^{-1}(\theta - m_k)(\theta - m_k)^TC_k^{-1} - C_k^{-1} \Bigr)\otimes\Bigl( C_k^{-1}(\theta - m_k)(\theta - m_k)^TC_k^{-1} - C_k^{-1} \Bigr) \dd\theta. 
\end{split}
\end{equation} 
{It is worth noting that \cref{eq:FI-C-approximation} is a 4th order tensor. To gain a more detailed understanding of this term, let us denote}
% \yc{(notation is not great)}
\begin{equation}
\begin{split}
    X_k :=& \frac{1}{4}\int \N_k \Bigl( C_k^{-1}(\theta - m_k)(\theta - m_k)^TC_k^{-1} - C_k^{-1} \Bigr)\otimes\Bigl( C_k^{-1}(\theta - m_k)(\theta - m_k)^TC_k^{-1} - C_k^{-1} \Bigr) \dd\theta
    \\
     =& \frac{1}{4}\int \N(y;0,I) C_k^{-1/2}\Bigl( yy^T - \I \Bigr)C_k^{-1/2}\otimes C_k^{-1/2}\Bigl( yy^T - \I \Bigr)C_k^{-1/2} \dd y\\& \qquad \text{ where } y = C_k^{-1/2}(\theta - m_k).
\end{split}
\end{equation} 
We can show that $X_k$ satisfies \cref{eq:FI-X}. To do so note that the $(ij,lm)$ entry of $X_k$ has the form
\begin{align*}
X_k[ij, lm] =& \frac{1}{4}\sum_{r,s,p,q}C_k^{-1/2}[i,r]C_k^{-1/2}[j,s]C_k^{-1/2}[l,p]C_k^{-1/2}[m,q]\int (y_ry_s-\delta_{r,s})(y_py_q-\delta_{p,q})\N(y;0,I) \dd y 
\\
=& \frac{1}{4}\sum_{r,s,p,q}C_k^{-1/2}[i,r]C_k^{-1/2}[j,s]C_k^{-1/2}[l,p]C_k^{-1/2}[m,q]\bigl(\delta_{r,p}\delta_{s,q} + \delta_{r,q}\delta_{s,p} \bigr) 
\\
=& \frac{1}{4} \Bigl(C_k^{-1}[i,l]C_k^{-1}[j,m] + C_k^{-1}[i,m]C_k^{-1}[j,l]\Bigr).
\end{align*}
Therefore,
\begin{equation}
\begin{aligned}
    (X_k Y)_{ij} &= \sum_{l,m} X_k[ij, lm] Y[l,m]\\
    & = \frac{1}{4} \sum_{l,m}\Bigl(C_k^{-1}[i,l]Y[l,m]C_k^{-1}[j,m] + C_k^{-1}[i,m]Y[l,m] C_k^{-1}[j,l]\Bigr)\\
    & = \frac{1}{4}(C_k^{-1}YC_k^{-1}+C_k^{-1}Y^TC_k^{-1})_{ij}.
\end{aligned}
\end{equation}
The proof is complete.

\section{Two-Dimensional Four-modal Problem}
\label{appendix-Two-Dimensional-Four-modal}

\begin{figure}[ht]
\centering
    \includegraphics[width=0.98\textwidth]{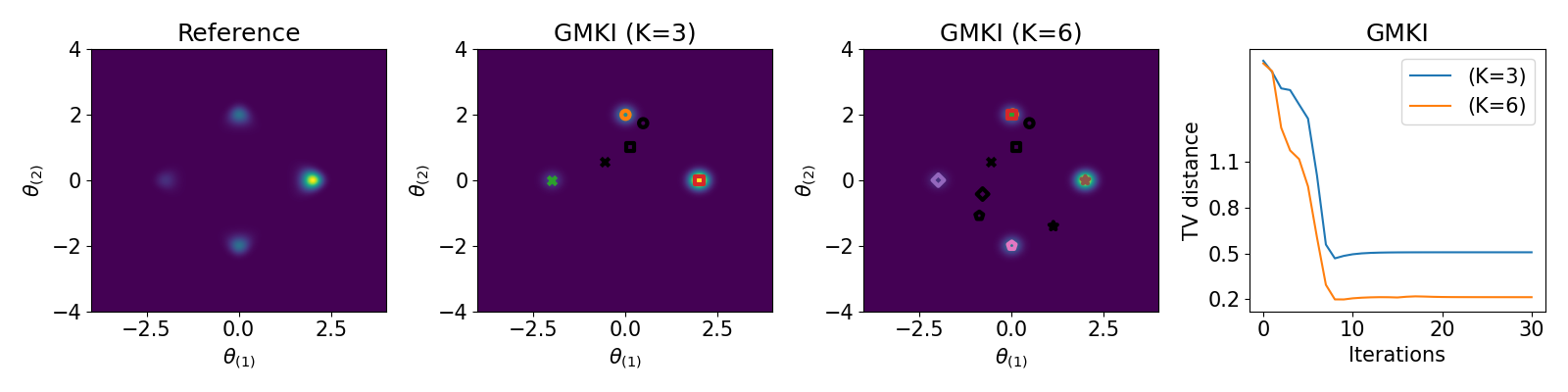}
   \caption{\dzh{Two-dimensional four-modal problem. 
    From left to right: reference posterior distribution (left), posterior distributions estimated by 3 and 6-modal GMKI (middle) at the 30th iteration (means $m_k$ (colored) and initial means (black) are marked), and total variation distance between the reference posterior distribution and the posterior distributions estimated by the GMKI (right).} }
    \label{fig:2D-density-four-modal}
\end{figure}

In this subsection, we consider a 2D four-modal inverse problem, associated with the forward model
$$y = \G(\theta) + \eta \quad \textrm { with } \quad 
y = \begin{bmatrix}
    4.2297\\
    4.2297
\end{bmatrix}, \quad \G(\theta) = 
\begin{bmatrix} 
(\theta_{(1)} - \theta_{(2)})^2 \\
(\theta_{(1)} + \theta_{(2)})^2
\end{bmatrix}.$$
Here $\theta = [\theta_{(1)}, \theta_{(2)}]^T$. We assume the noise distribution is $\eta \sim \N(0, I)$ and consider the prior distribution $\rho_{\rm prior} \sim \N([0.5,0]^T, I)$.
%%%%%%%%%%
The reference posterior distribution has four modes with varying weights. 
We apply GMKI with $K=3$ and $6$ modes,  randomly initialized based on the prior distribution, and assign equal weights to these components. The estimated posterior distributions obtained by GMKI at the 30th iteration, along with the convergence in terms of total variation distance, are shown in \cref{fig:2D-density-four-modal}.  When $K=3$ only three target modes are captured; when $K=6$, all target modes are captured, and the approximation error becomes significantly smaller.

\section{Limitation of the GMKI}
\label{appendix-GMKI-low-dim-manifold-posterior}
\begin{figure}[ht]
\centering
    \includegraphics[width=0.98\textwidth]{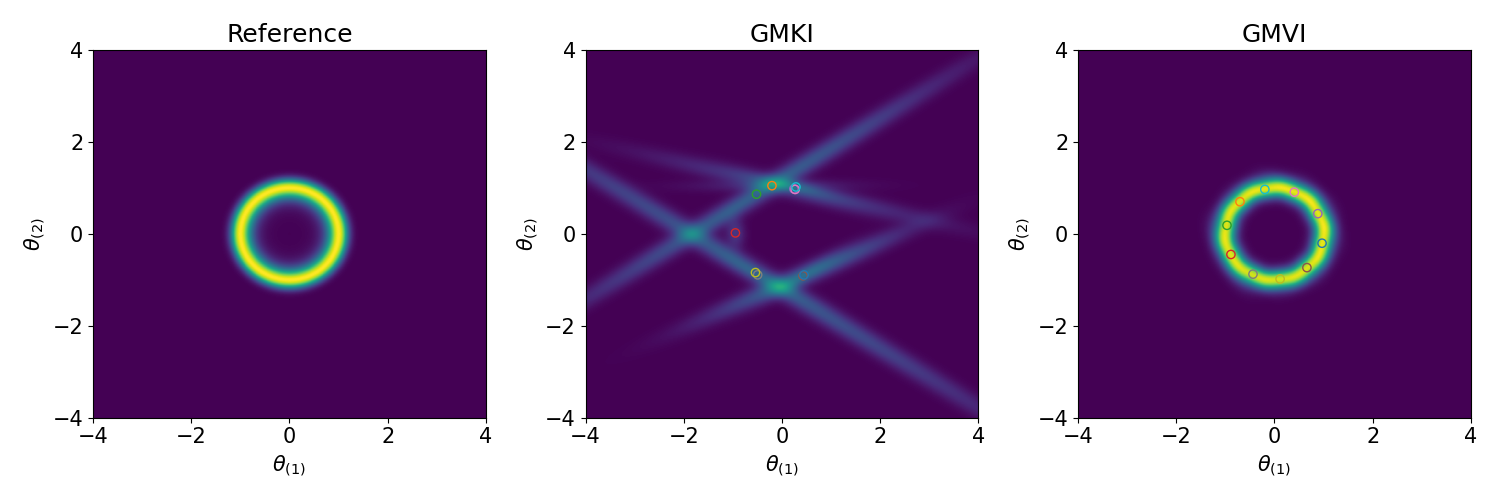}
    \caption{Circle shape posterior: reference posterior distribution (left), posterior distribution obtained by 10-modal GMKI (middle), posterior distribution obtained by 10-modal GMVI (right). Means $m_k$ of Gaussian components are marked. In the middle, the color appears lighter because the distribution is concentrated along a thin, elongated line, resulting in a visually diluted effect.}
    \label{fig:2D-circle-shape-density}
\end{figure}

% As depicted in \cref{ssec:1d-bimodal}, increasing the number of Gaussian components for GMKI may not necessarily lead to an improved approximation.
There can be multimodal problems with many modes concentrating on a low dimensional manifold. 
GMKI may fail in such a case.
To illustrate, we consider the posterior distribution $\exp(-\Phi_R(\theta))$ in $\mathbb{R}^2$, where $\Phi_R(\theta) = \frac{(1 - \theta_{(1)}^2 - \theta_{(2)}^2)^2}{2\sigma_{\eta}^2}$ and $\sigma_{\eta} = 0.3$. Clearly the mass is distributed along the unit circle, as depicted on the left side of \cref{fig:2D-circle-shape-density}. We sample this density with GMKI and Gaussian mixture variational inference (GMVI), described in \cref{eq:Appr-FR-GM}. Note that GMKI can be seen as a derivative free approximation of GMVI so this study is for the purpose of understanding the effect of the derivative free approximation step. We and initialize both methods with $10$ Gaussian components with means randomly sampled from $\N(0,I)$ and the same identity covariance $I$.

The sampling results obtained by GMKI at the 30th iteration are presented in the middle of \cref{fig:2D-circle-shape-density}. 
While the means of these Gaussian components migrate towards the unit circle, the covariance associated with each Gaussian component are elongated in the tangent direction. Consequently, the overall approximation of the target distribution is inaccurate. The covariance of these Gaussian components bear resemblance to those seen in the Laplace approximation.
Indeed, the Laplace approximation at any maximum a posteriori (MAP) of the target distribution has the form
$\N(\theta; m , H^\dagger)$; here, $m = [m_{(1)},  m_{(2)}]$ lies on the unit circle and  
\begin{equation}
    H = -\nabla_{\theta}\nabla_{\theta}\Phi_R(\theta)\vert_{\theta = m} = \frac{4}{\sigma_{\eta}^2}\begin{bmatrix}
   m_{(1)}^{(2)} & m_{(1)}m_{(2)}\\
   m_{(1)}m_{(2)} & m_{(2)}^{2} 
\end{bmatrix}.
\end{equation}
It is worth mentioning that $H$ exhibits singularity, particularly along the tangent direction of the unit circle. Hence the Laplace approximation is degenerate and concentrates on the tangential line of the unit circle at $m$.

To further explore this issue, we turn to use the algorithm for GMVI. This approach requires the evaluation of the gradient and Hessian of $\log \rho_{\rm post}$.
We approximate these Gaussian integrations in \cref{eq:Appr-FR-GM} using the modified unscented transform, as detailed in \cite[Eq. 37]{huang2022efficient}. Moreover we employ a time-step of $\Delta t = 0.01$, which leads to a stable numerical scheme in our implementation. The outcome of GMVI at the 1000th iteration is depicted in the right of \cref{fig:2D-circle-shape-density}. The result matches the reference well.
Since the main difference between GMVI \cref{eq:Appr-FR-GM} and the continuous time dynamics of GMKI is the derivative-free Kalman approximation for the gradient terms $\nabla_{\theta}\log \rho_{\rm post}$ and $\nabla_{\theta}\nabla_{\theta}\log \rho_{\rm post}$, we understand that the Kalman approximation step leads to the failure of GMKI for sampling the above distribution. It is the goal of future study to investigate other derivative free approximations that can circumvent this failure. 

\section*{References}
\bibliographystyle{unsrt}
\bibliography{references}
\end{document}